\pdfoutput=1

\documentclass[11pt]{article}

\usepackage{ACL2023}

\usepackage{times}
\usepackage{latexsym}

\usepackage[T1]{fontenc}

\usepackage[utf8]{inputenc}

\usepackage{microtype}

\usepackage{inconsolata}

\usepackage{graphicx}
\usepackage{subfig}
\usepackage{booktabs}

\usepackage{colortbl}
\definecolor{mygray}{gray}{.9}
\usepackage{algorithm}
\usepackage{algorithmic}
\usepackage{color}
\usepackage{multirow}
\usepackage{amsfonts,amssymb}
\usepackage{amsmath}
\usepackage{booktabs}
\usepackage{mathtools}
\usepackage{url}
\usepackage{stfloats}
\usepackage{courier}

\title{Learning "O" Helps for Learning More: \\Handling the Unlabeled Entity Problem for Class-incremental NER}

\author{
    Ruotian Ma\textsuperscript{\rm 1}\thanks{\ \ Equal contribution.} , Xuanting Chen\textsuperscript{\rm 1}\footnotemark[1] ,
    Lin Zhang\textsuperscript{\rm 1}, 
    Xin Zhou\textsuperscript{\rm 1},
     \\ 
    \textbf{
    Junzhe Wang\textsuperscript{\rm 1},
    Tao Gui\textsuperscript{\rm 2}\thanks{\ \  Corresponding authors.}\ ,
    Qi Zhang\textsuperscript{\rm 1}\footnotemark[2] ,
    Xiang Gao\textsuperscript{\rm 3},
    Yunwen Chen\textsuperscript{\rm 3}}\\
  $^1$School of Computer Science, Fudan University, Shanghai, China \\
  $^2$Institute of Modern Languages and Linguistics, Fudan University, Shanghai, China \\
  $^3$DataGrand Information Technology (Shanghai) Co., Ltd.\\
  \texttt{\{rtma19,xuantingchen21,tgui,qz\}@fudan.edu.cn}
}

\begin{document}
\maketitle
\begin{abstract}
As the categories of named entities rapidly increase, the deployed NER models are required to keep updating toward recognizing more entity types, creating a demand for class-incremental learning for NER. Considering the privacy concerns and storage constraints, the standard paradigm for class-incremental NER updates the models with training data only annotated with the new classes, yet the entities from other entity classes are unlabeled, regarded as "Non-entity" (or "O"). In this work, we conduct an empirical study on the "Unlabeled Entity Problem" and find that it leads to severe confusion between "O" and entities, decreasing class discrimination of old classes and declining the model's ability to learn new classes. To solve the Unlabeled Entity Problem, we propose a novel representation learning method to learn discriminative representations for the entity classes and "O". Specifically, we propose an entity-aware contrastive learning method that adaptively detects entity clusters in "O". Furthermore, we propose two effective distance-based relabeling strategies for better learning the old classes. We introduce a more realistic and challenging benchmark for class-incremental NER, and the proposed method achieves up to 10.62\% improvement over the baseline methods.

\end{abstract}
\section{Introduction}\label{sec:intro}

Existing Named Entity Recognition systems are typically trained on a large-scale dataset with pre-defined entity classes, then deployed for entity recognition on the test data without further adaptation or refinement \cite{li2020survey, wang-etal-2022-shot,liu2021crossner,ijcai2022p595}. In practice, the newly-arriving test data may include new entity classes, and the user's required entity class set might keep expanding. Therefore, it is in demand that the NER model can be incrementally updated for recognizing new entity classes. However, one challenge is that the training data of old entity classes may not be available due to privacy concerns or memory limitations \cite{li2017learning,zhang2020class}. Also, it is expensive and time-consuming to re-annotate all the old entity classes whenever we update the model \cite{MatthiasDelange2021ACL, bang2021rainbow}. To solve the problem, \citet{monaikul2021continual} proposes to incrementally update the model with new datasets only covering the new entity classes, adopted by following studies as standard \textbf{class-incremental NER} paradigm.

\begin{figure}
    \centering
    \includegraphics[width=1.0\linewidth]{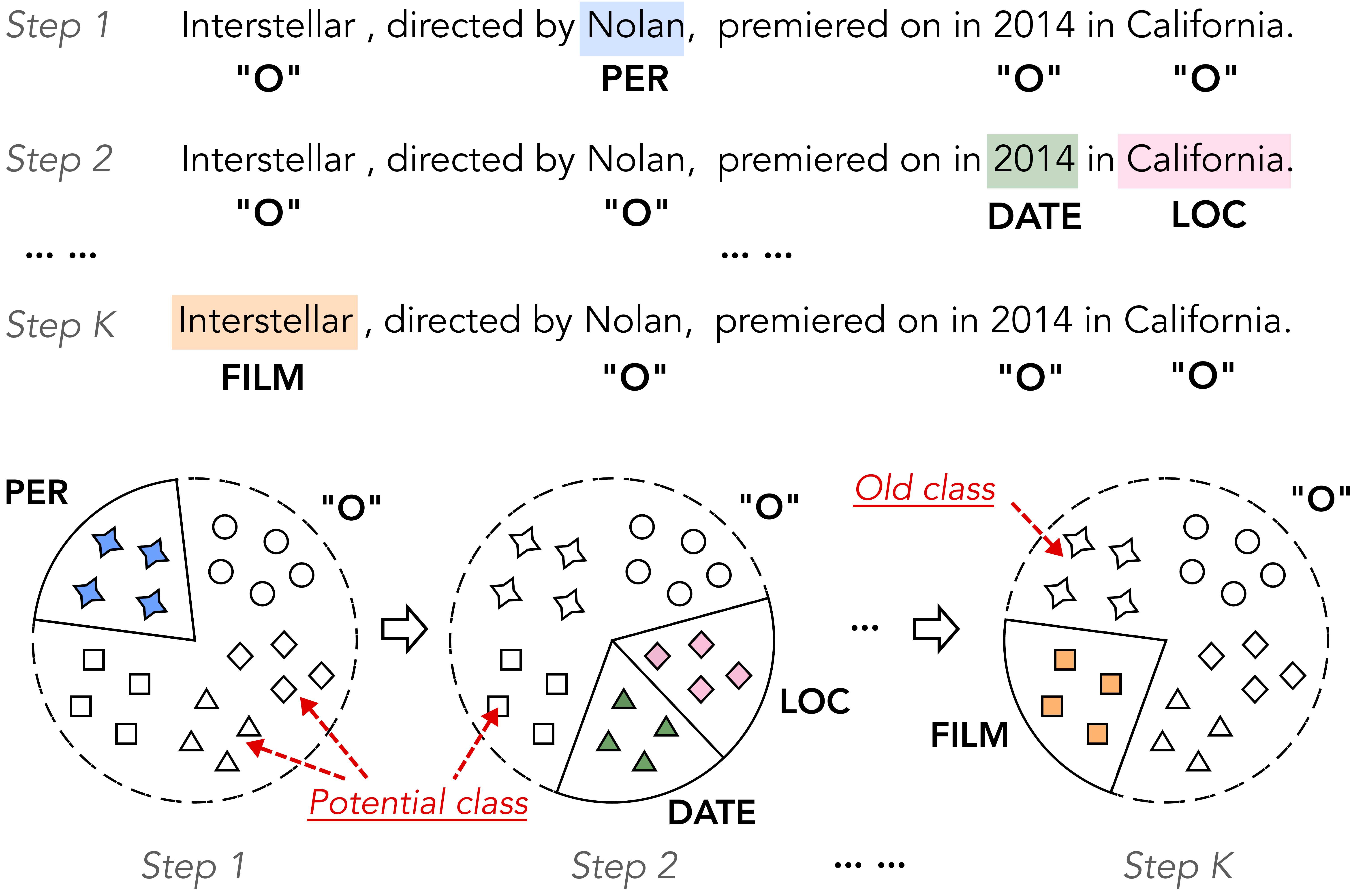}
    \caption{Problems of class-incremental NER. In each incremental step, the data is only labeled with current classes, so the "O" class actually contains entities from old classes and entities from potential classes.}
    \label{problem}
    \vspace{-0.4cm}
\end{figure}

However, as NER is a sequence labeling task, annotating only the new classes means entities from other entity classes are regarded as "Non-entity" (or "O") in the dataset. For example, in step 2 in Fig.1, the training data for model updating is only annotated with "LOC" and "DATE", while the entities from "PER" and "FILM" are unlabeled and regarded as "O" during training. We refer to this problem as the "Unlabeled Entity Problem" in class-incremental NER, which includes two types of unlabeled entities: (1) old entity classes (e.g., "PER" in step 2) that the model learned in previous steps are unlabeled in the current step, causing the model catastrophically forgetting these old classes. \cite{ lopez2017gradient, castro2018end} (2) potential entity classes that are not annotated till the current step, yet might be required in a future step. For example, the "FILM" class is not annotated till step 2, yet is required in step K.

In this work, we conduct an empirical study to demonstrate the significance of the "Unlabeled Entity Problem" on class-incremental NER. We observe that: (1) The majority of prediction errors come from the confusion between entities and "O". (2) Mislabeled as "O" leads to the reduction of class discrimination of old entities during incremental learning. (3) The model's ability to learn new classes also declines as the potential classes are unlabeled during incremental training. These problems attribute to the serious performance drop of incremental learning with the steps increasing.

To tackle the Unlabeled Entity Problem, we propose a novel representation learning method for learning discriminative representations for the unlabeled entity classes and "O". Specifically, we propose an entity-aware contrastive learning approach, which adaptively detects entity clusters from "O" and learns discriminative representations for these entity clusters. To further maintain the class discrimination of old classes, we propose two distance-based relabeling strategies. By relabeling the entities from old classes with high accuracy, this practice not only keeps the performance of old classes, but also benefits the model's ability to separate new classes from "O".

We also argue that the experimental setting of previous works \citet{monaikul2021continual} is less realistic. Specifically, they introduce only one or two entity classes in each incremental step, and the number of total steps is limited. In real-world applications, it is more common that a set of new categories is introduced in each step (e.g., a set of product types), and the incremental learning steps can keep increasing. In this work, we provide a more realistic and challenging benchmark based on the Few-NERD dataset \cite{ding-etal-2021-nerd},  following the settings of previous studies \cite{rebuffi2017icarl,li2017learning}. We conduct intensive experiments on the proposed methods and other comparable baselines, verifying the effectiveness of the proposed method \footnote{Our code is publicly available at \url{https://github.com/rtmaww/O_CILNER}. }.

To summarize the contribution of this work:
\begin{itemize}
\setlength{\itemindent}{0em}
\setlength{\itemsep}{0em}
\setlength{\topsep}{-2.0em}
    \item We conduct an empirical study to demonstrate the significance of the "Unlabeled Entity Problem" in class-incremental NER.
    \item Based on our observations, we propose a novel representation learning approach for better learning the unlabeled entities and "O", and verify the effectiveness of our method with extensive experiments.
    \item We provide a more realistic and challenging benchmark for class-incremental NER.
\end{itemize}

\section{Class-incremental NER}
In this work, we focus on class-incremental learning on NER. Formally, there are $N$ incremental steps, corresponding to a series of tasks $\{\mathcal{T}_1, \mathcal{T}_2, \dots, \mathcal{T}_N \}$. 
Here, $\mathcal{T}_t=(\mathcal{D}_t^{tr}, \mathcal{D}_t^{dev}, \mathcal{D}_t^{test}, \mathcal{C}_{t,new}, \mathcal{C}_{t,old}  )$ is the task at the $t^{th}$ step.
$\mathcal{C}_{t,new}$ is the label set of the current task, containing only the \textbf{new classes} introduced in the current step (e.g., \{"LOC", "DATE"\} in Fig.1, step 2). $\mathcal{C}_{t,old}=\bigcup\limits_{i=1}^{t-1}\mathcal{C}_{i,new} \cup \{``O"\}$ is the label set of \textbf{old classes}, containing all classes in previous tasks and the class "O" (e.g., \{"PER", "O"\} in Fig.1, step 2).
$\mathcal{D}_t^{tr}=\{{X}^j_t,{Y}^j_t\}^n_{j=1}$ is the training set of task $t$, where each sentence $X_t^j=\{x_t^{j,1}, \dots, x_t^{j,l}\}$ and $Y_t^j=\{y_t^{j,1}, \dots, y_t^{j,l}\}, y_t^{j,k} \in \mathcal{C}_{t,new}$ is annotated with only the new classes. In each step $t$, the model $\mathcal{A}_{t-1}$ from the last step needs to be updated with only the data $\mathcal{D}_{t}^{tr}$ from the current step, and is expected to perform well on the test set covering all learnt entity types $\mathcal{C}_t^{all}=\mathcal{C}_{t,new} \cup \mathcal{C}_{t,old}$.

\section{The Importance of Unlabeled Entity Problem in Class-incremental NER}
In this section, we demonstrate the importance of the Unlabeled Entity Problem in Class-incremental NER with empirical studies.
We conduct experiments on a challenging dataset, the Few-NERD dataset, to investigate the problems in class-incremental NER. We conduct experiments with two existing methods: (1) \textbf{iCaRL} \cite{rebuffi2017icarl}, a typical and well-performed method in class-incremental image classification. (2) \textbf{Continual NER} \cite{monaikul2021continual}, the previous state-of-the-art method in class-incremental NER. More details of the dataset and the baseline methods can be found in Section \ref{experiment}. 

\noindent \textbf{Observation 1: The majority of prediction errors come from the confusion between entities and "O". }
In Fig.2, we show the distributions of prediction errors of different models in step 6, where the y-axis denotes samples belonging to "O" or the classes of different tasks. The x-axis denotes the samples are wrongly predicted as "O" or as classes from different tasks. Each number in a grid denotes the number of error predictions. From the results, we can see that the majority of error predictions are samples belonging to "O" wrongly predicted as entities (the first row of each model), indicating serious confusion between "O" and entity classes, especially the old entity classes. 
As explained in Section \ref{sec:intro}, the training data of each new task is only annotated with the new entity classes and the entities from old classes are labeled as "O". As the training proceeds, the class variance between 
 the true "O" and old entity classes will decrease, leading to serious confusion of their representations.

\begin{figure}
    \centering
    
\centering
\subfloat[iCaRL]{\includegraphics[width=0.5\linewidth]{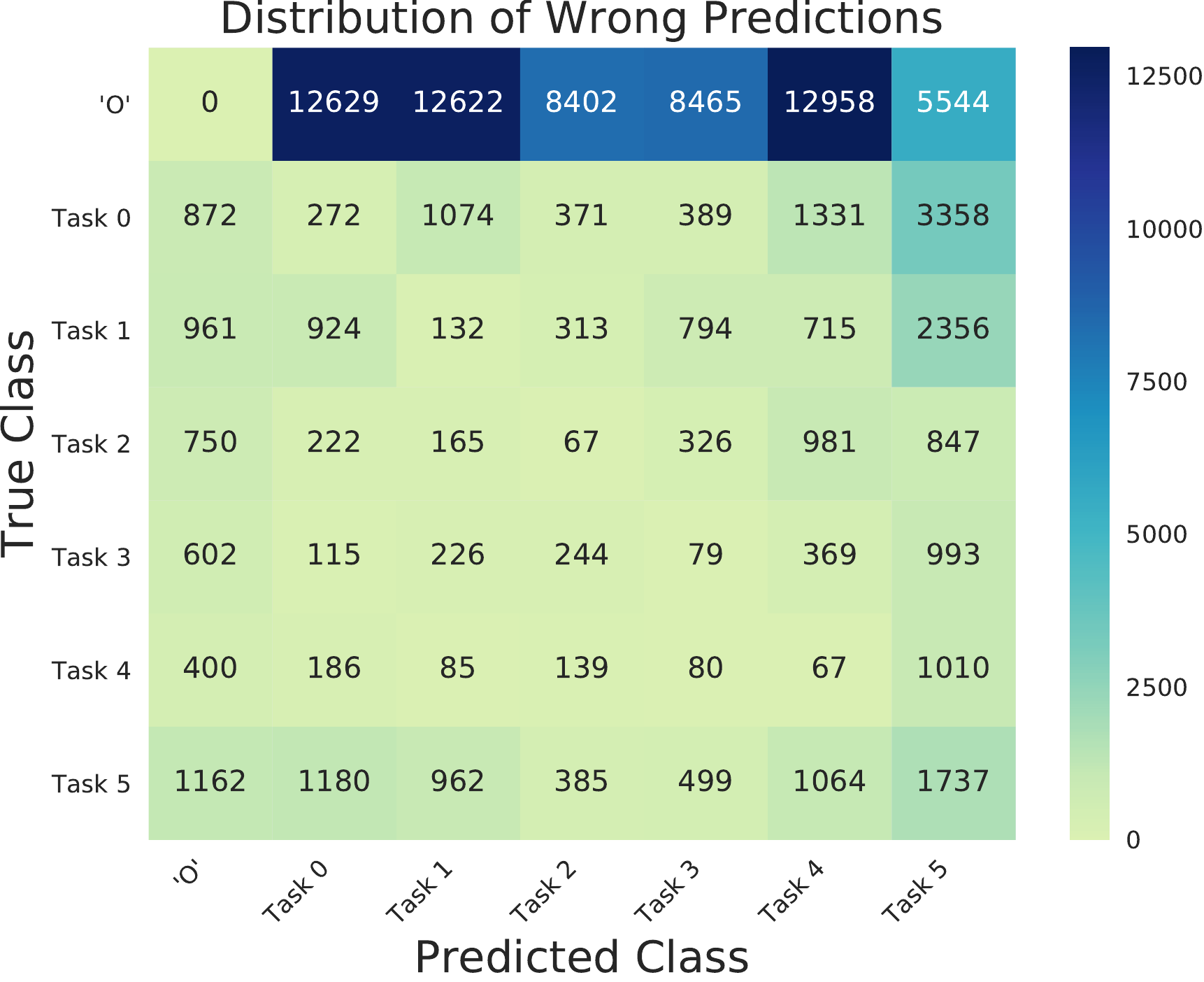}}
\subfloat[Continual NER]{\includegraphics[width=0.5\linewidth]{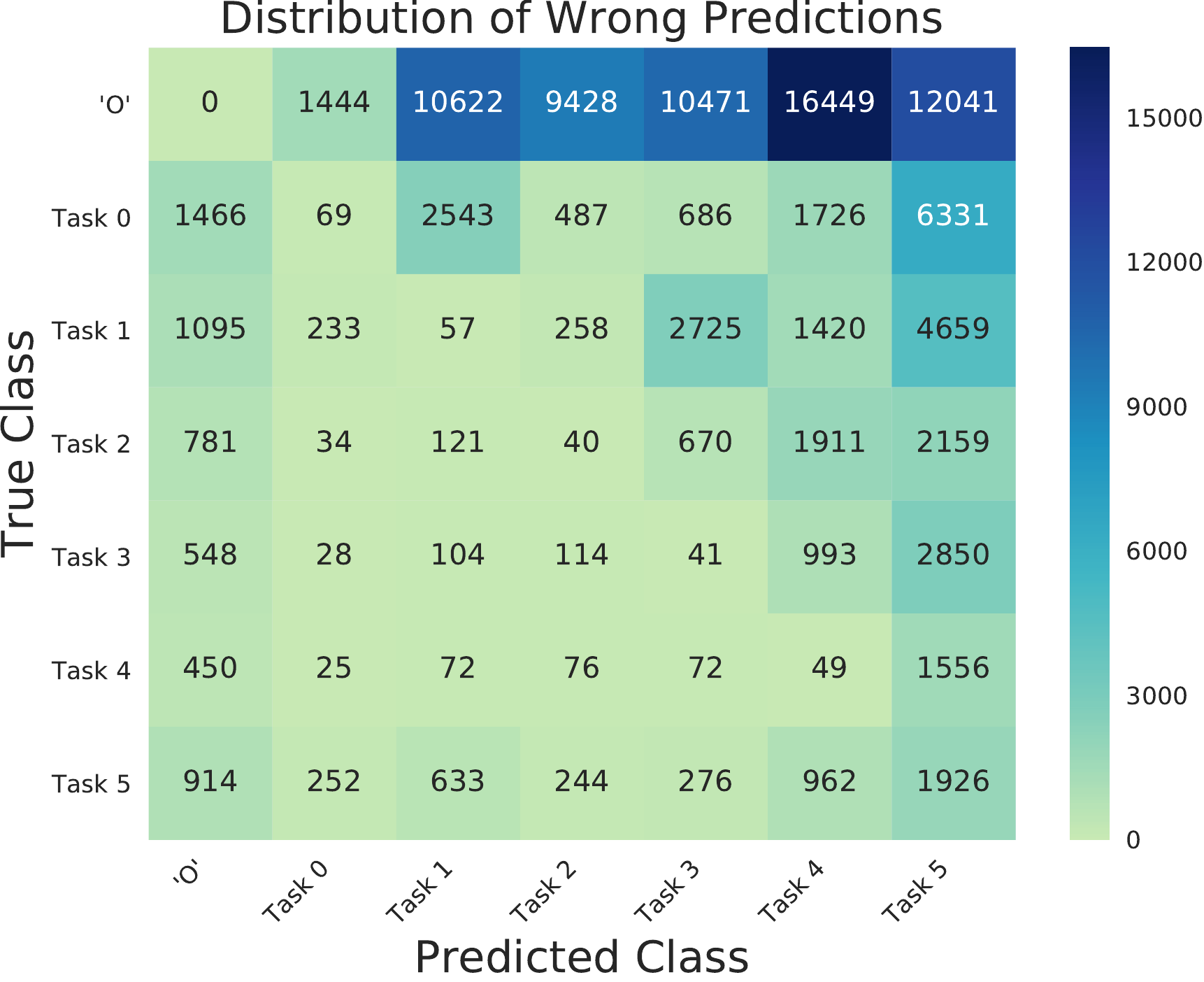}} \\
    \caption{Distributions of prediction errors of different models in step 6. The first row represents the number of samples belonging to the "O" class wrongly recognized as entities, which  shows the severe confusion between "O" and entity classes.}
    \label{wrongpred}
    \vspace{-0.3cm}
\end{figure}

\begin{figure}
    
\subfloat[Continual NER, step 2]{\includegraphics[width=0.5\linewidth]{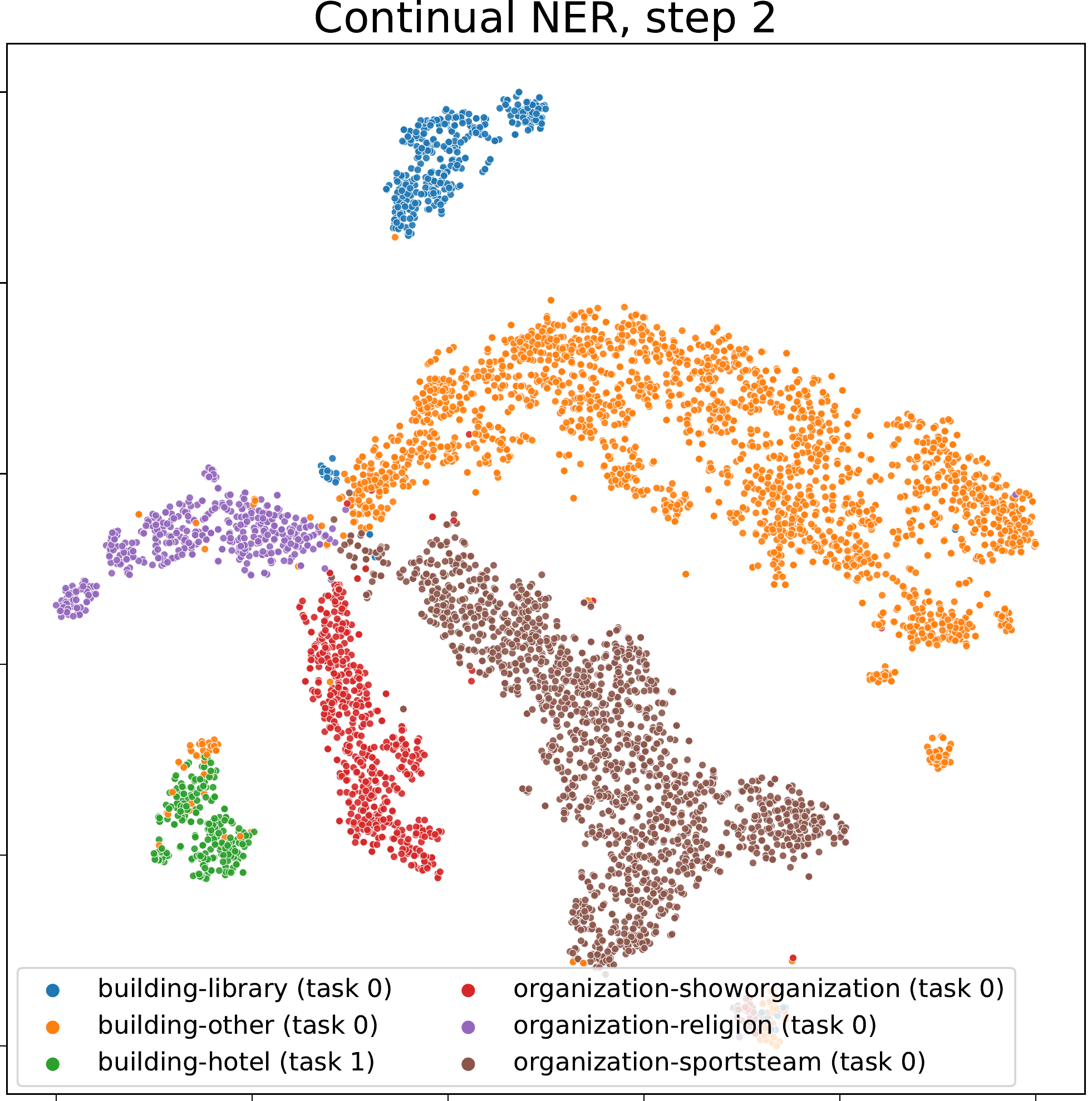}}
\subfloat[Continual NER, step 5]
{\includegraphics[width=0.5\linewidth]{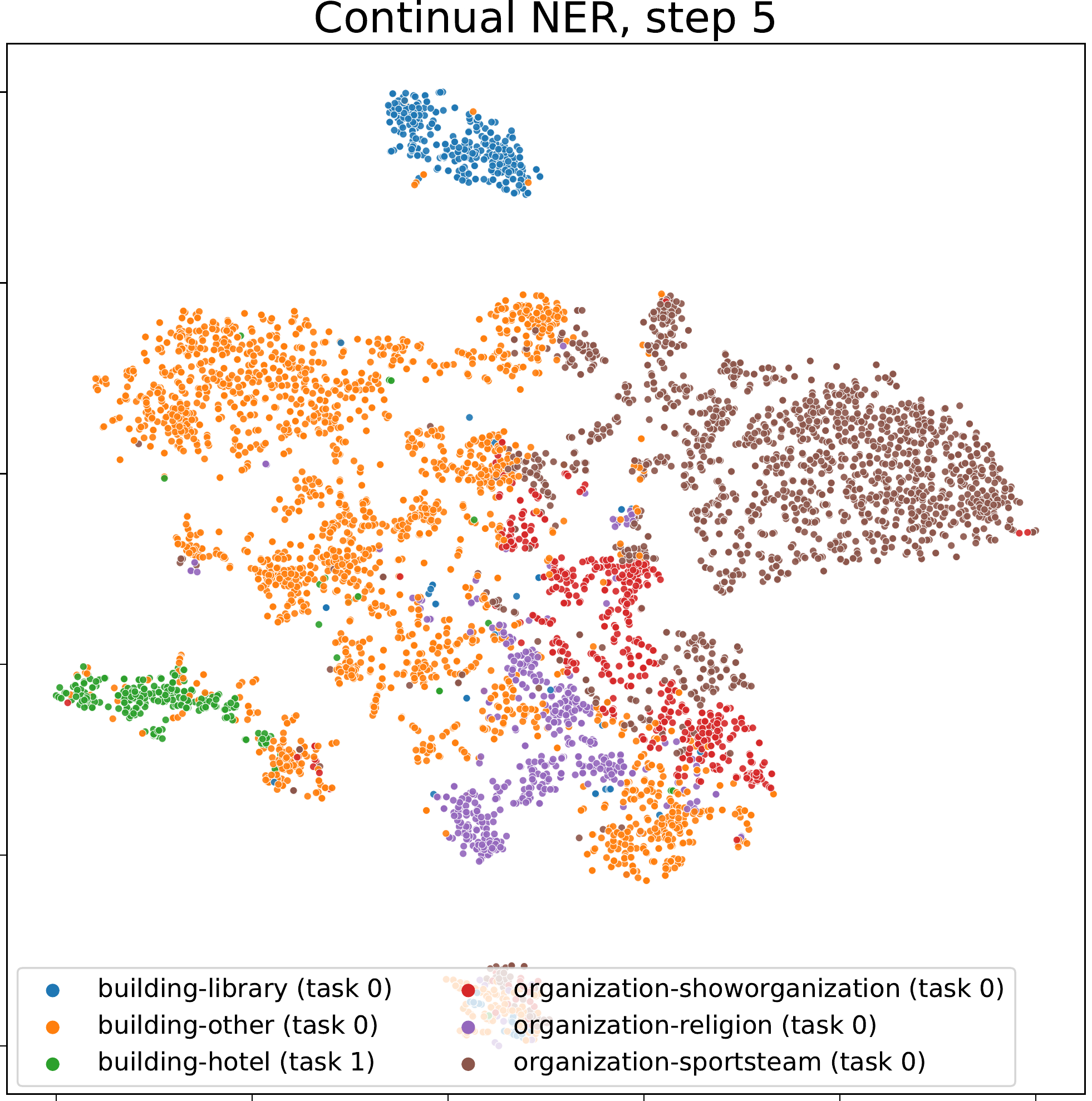}} \\

    \caption{Visualization of the representation variation of the old classes during incremental learning. The class discrimination seriously decrease in step 5.}
    \label{tsne}
    \vspace{-0.2cm}
\end{figure}

\paragraph{Observation 2: Old entity classes become less discriminative during incremental learning.}
We further investigate the representation variation of old classes during incremental learning. As shown in Fig.3, we select similar classes from step 0 and step 1, and visualize their representations after step 2 and step 5. The results show that the representations of these classes are discriminative enough in step 2. However, after a series of incremental steps, the representations of these old classes become less discriminative, leading to decreasing performance of old classes. This phenomenon also indicates the influence of the unlabeled entity problem on the unlabeled old classes.

\begin{table}[h]
\setlength\tabcolsep{3.5pt}
\centering
\small

\begin{tabular}{lccccccc}
\toprule
\textbf{Steps} & \textbf{0} & \textbf{1} & \textbf{2} & \textbf{3} & \textbf{4} & \textbf{5}& \textbf{6} \\
\midrule
\bfseries Full Data &72.7 & 69.2 & 68.3 & 67.0 & 67.3 & 69.1 & 68.8\\
\bfseries iCaRL & 71.3 & 56.9 & 52.6 & 48.8 & 53.4 & 48.1 & 39.6 \\
\bfseries Con. NER  & 72.4 & 63.5 & 56.9 & 52.5 & 56.8 & 51.8 & 42.2 \\
\bottomrule

\end{tabular}
\caption{Performance of the new classes on dev set (only containing the new classes) keep decreasing during incremental learning. Here, Full Data is the model trained with datasets labeled with both old and new classes. }
\label{dev}
\vspace{-0.2cm}
\end{table}

\paragraph{Observation 3: The model's ability to learn new classes declines during incremental learning.} 
Finally, we conduct an experiment to investigate the model's ability to learn new classes. In Table \ref{dev}, we test the results of new classes in each step on dev sets that only contain these new classes. Here, \textbf{Full Data} is a baseline that trains on datasets that both old and new classes are annotated. 
Surprisingly, we find that the performance of the new classes of {iCaRL} and {Continual NER} keeps decreasing during incremental learning, compared to the stable performance of {Full Data}.
This phenomenon is also related to the Unlabeled Entity Problem. As explained in the introduction, the potential entity classes (i.e., the entity classes that might be needed in a future step) are also unlabeled and regarded as "O" during incremental learning. As a result, the representations of these classes become less separable from similar old classes (also labeled as "O"), thus hindering the model's ability to learn new classes.

\paragraph{Conclusion to the Observations:} Based on above observations, we propose that appropriate representation learning are required to tackle the Unlabeled Entity Problems. The representations of entity and "O" are expected to meet the following requirements: (1) The "O" representations are expected to be distinct from the entity representations, so as to decline the confusion between "O" and entities (\textbf{Observation 1}).  (2) The representations of old entity classes are expected to keep discriminative in spite of being labeled as "O" (\textbf{Observation 2}). (3) The potential entity class are expected to be detected and separated from "O", and also be discriminative to other entity classes (\textbf{Observation 3}). 
These observations and conclusions contribute to the motivation of the proposed method.

\section{Handling the Unlabeled Entity Problem}

In order to learn discriminative representations for unlabeled entity classes and the true "O" (connected to \textbf{Observations 1, 2, 3}), we propose entity-aware contrastive learning, which adaptively detects entity clusters in "O" during contrastive learning. To further maintain the class discrimination of old classes (connected to \textbf{Observation 2}), we propose two distance-based relabeling strategies to relabel the unlabeled entities from old classes in "O". Additionally, we propose the use of the Nearest Class Mean classifier based on learnt representations in order to avoid the prediction bias of linear classifier.

\noindent \textbf{Rehearsal-based task formulation}
To better learn representations for entities and "O", in this work, we follow the memory replay (rehearsal) setting adopted by most of the previous works \cite{rebuffi2017icarl, mai2021supervised,verwimp2021rehearsal}. Formally, we retain a set of exemplars $\mathcal{M}_c=\{{x}^i_c,{y}^i_c, \overline{X}^i_c\}^K_{i=1}$ for each class $c$, where ${x}^i_c$ refers to one token $x$ labeled as class $c$ and $\overline{X}$ is the context of $x$ labeled as "O". In all our experiments, we set $K=5$ \footnote{We set a small $K$ because the class number can be large.}.

\begin{figure*}
    \centering
    \includegraphics[width=0.9\linewidth]{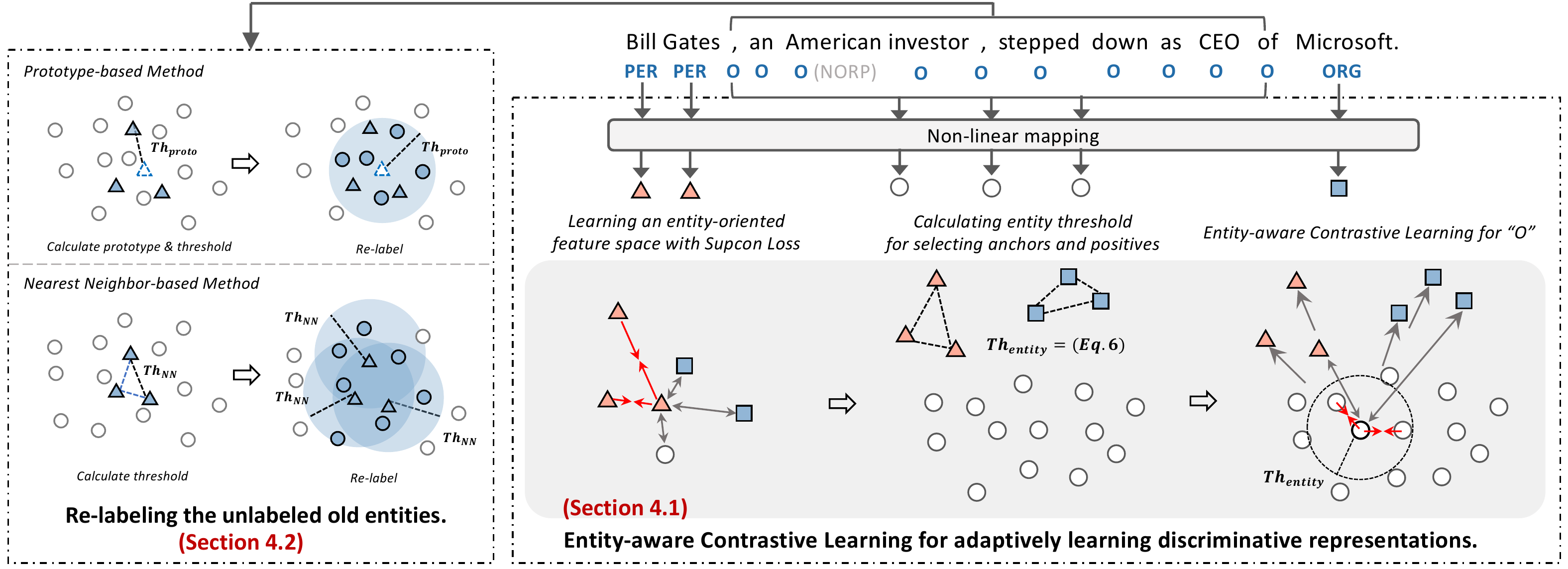}    \caption{Overview of the proposed representation learning method: (1) We propose an entity-aware contrastive learning method to adaptively detect entity clusters from "O" and learn discriminative representations for these entities. (2) We propose two distance-based relabeling strategies to further maintain the performance of old classes. }
    \label{fig:example}
    \vspace{-0.2cm}
\end{figure*}

\subsection{Entity-aware Contrastive Learning}
In this section, we introduce the entity-aware contrastive learning, which dynamically learns entity clusters in "O". To this aim, we first learn an entity-oriented feature space, where the representations of entities are distinctive from "O". This entity-oriented feature space is learnt through contrastive learning on the labeled entity classes in the first $M$ epochs of each step. Based on the entity-oriented feature space, we further conduct contrastive learning on "O", with the anchors and positive samples dynamically selected based on an entity threshold.

\paragraph{Learning an Entity-oriented Feature Space.} 
Firstly, we are to learn an entity-oriented feature space, where the distance between representations reflects entity semantic similarity, i.e., representations from the same entity class have higher similarity while keeping the distance from other classes. This feature space is realized by learning a non-linear mapping $F(\cdot)$ on the output representations $\mathbf{h}$ of PLM. We adopt cosine similarity as the similarity metric and train with the Supervised Contrastive Loss \cite{NEURIPS2020_d89a66c7}:

\begin{equation}\label{supcon}
\setlength\abovedisplayskip{0pt}
\setlength\belowdisplayskip{0pt}
\begin{split}
    L_{SCL} = \sum_{i\in I} \frac{-1}{|P(i)|} \sum_{p\in P(i)} log \frac{e^{{s}(\mathbf{z}_i, \mathbf{z}_p)/\tau}}{\sum_{a\in A(i)}e^{\mathbf{s}(\mathbf{z}_i,\mathbf{z}_a)/\tau}}
    \end{split}
\end{equation}
where $\mathbf{z}=F(\mathbf{h})$ denotes the representation after the mapping and ${s}(\cdot)$ is the cosine similarity. 

Here, we apply contrastive learning only on the entity classes, thus we define:
\begin{equation}\label{anchor_define}
\begin{split}
    &I =\{i\mid i \in Index(\mathcal{D}_t^{tr}), \; y_i \neq ``O"\} \\
    &A(i) = \{j\mid j \in Index(\mathcal{D}_t^{tr}), \; j \neq i\} \\
    &P(i)=\{p\mid p\in A(i),\; y_p=y_i\}
    \end{split}
\end{equation}
where the anchor set $I$ \textbf{only includes entity tokens}.
We train with $L_{SCL}$ in the first $K$ epochs, improving the representations of entities and obtaining an entity-oriented feature space.

\paragraph{Calculating an entity threshold for anchors and positive samples selection.}
Based on the entity-oriented feature space, we are to dynamically select possible entity clusters in "O" and further optimize their representations via contrastive learning. This selection is realized by a dynamically adjusted \textit{entity threshold}.

Specifically, we first define the \textit{class similarity} $S_c$ as the average of exemplar similarities inside each class:
\begin{equation}
\small
\setlength\abovedisplayskip{5pt}
\setlength\belowdisplayskip{5pt}
\begin{split}
    S_c = \frac{1}{|\mathcal{M}_c|}\sum_{x_i, x_j \in \mathcal{M}_c, \atop x_i \neq x_j} s(F(h(x_i)), F(h(x_j)))
    \end{split}
\end{equation}

Then, we sort the \textit{class similarity} of all classes and choose the median as the \textit{entity threshold} $\mathcal{T}_{ent}$ (here we simply choose the median for a modest threshold):
\begin{equation}
\small
\setlength\abovedisplayskip{5pt}
\setlength\belowdisplayskip{5pt}
    \mathcal{T}_{ent} = Sorted(\{S_1, \dots, S_{|\mathcal{C}_t^{all}|}\})[i], i=\frac{|\mathcal{C}_t^{all}|}{2}
\end{equation}
During contrastive learning for "O", we re-calculate $\mathcal{T}_{ent}$ before each epoch to dynamically adjust the threshold based on convergence degree.

\paragraph{Contrastive Learning for "O" with the entity threshold}
Based on entity threshold $\mathcal{T}_{ent}$, we then apply the entity-aware contrastive learning for "O" with auto-selected anchors and positive samples. Specifically, we re-define Eq.\ref{anchor_define} as:
\begin{equation}\label{o_def}
\small
\begin{split}
    &I_O = \{i \mid \exists j \neq i, y_j=y_i=``O", s(\mathbf{z}_i,\mathbf{z}_j)>\mathcal{T}_{ent}\}\\
    &P_O(i) = \{p \mid p \neq i, y_p=``O", s(\mathbf{z}_i,\mathbf{z}_p)>\mathcal{T}_{ent}\} \\
    &A_O(i) = P_O(i)\cup\{n\mid y_n\in \mathcal{C}_{t,new}\}
\end{split}
\end{equation}
Then, we define the entity-aware contrastive loss of "O" by adopting Eq.\ref{supcon} with the definition in Eq.\ref{o_def}:
\begin{equation}
\begin{split}
    L_{SCL,O} = L_{SCL}(I_O, P_O,A_O)
\end{split}
\end{equation}
In the last $N-K$ epochs, we jointly optimize the representations of entities and "O" by:
\begin{equation}
\setlength\abovedisplayskip{3pt}
\setlength\belowdisplayskip{0pt}
\begin{split}
    L = L_{SCL,O} + L_{SCL}
\end{split}
\end{equation}

\subsection{Relabeling Old Entity Classes}

In order to further retain the class discrimination of old classes, we propose two distance-based relabeling strategies to recognize and relabel the unlabeled old-class entities in "O". These two strategies are designed to make use of the previous model $\mathcal{A}_{t-1}$ and the exemplar set $\mathcal{M}$.

\paragraph{Relabeling with Prototypes.}
This strategy relabels "O" samples based on their distance to the class prototypes. Specifically, we first calculate the prototype of each class based on the representations of exemplars from the old model $\mathcal{A}_{t-1}$.
\begin{equation}\label{prototype}
\setlength\abovedisplayskip{3pt}
\setlength\belowdisplayskip{0pt}
\begin{split}
    \mathbf{p}_c = \frac{1}{|\mathcal{M}_c|}\sum_{x\in \mathcal{M}_c} h_{t-1}(x)
    \end{split}
\end{equation}
Then, we define a relabeling threshold, denoted as the \textit{prototype relabeling threshold}, by calculating the lowest similarity of all exemplars with their prototypes:
\begin{equation}
\setlength\abovedisplayskip{3pt}
\setlength\belowdisplayskip{0pt}
\begin{split}
    \mathcal{T}{h}_{proto}=\beta \cdot \min_{(x,y) \in \mathcal{M}_c  \atop c \in \mathcal{C}_{t,old}}\{s(h_{t-1}(x),\mathbf{p}_y)\}
\end{split}
\end{equation}
where $\beta$ is a hyper-parameter to control the relabeling degree. Next, for each "O" sample $x_i$ in $\mathcal{D}_t^{tr}$, we relabel it only if its highest similarity to prototypes is larger than $\mathcal{T}{h}_{proto}$:
\begin{equation}
\setlength\abovedisplayskip{3pt}
\setlength\belowdisplayskip{0pt}
\begin{split}
\mathcal{S} =& \{s(h_{t-1}(x_i), \mathbf{p}_c) \mid c \in \mathcal{C}_{t,old}\} \\
    y_i=&\mathop{\arg\max}_{c} \mathcal{S}, \quad
    if \max \mathcal{S} > \mathcal{T}{h}_{proto}
\end{split}
\end{equation}

\paragraph{Relabeling with Nearest Neighbors.} In this approach, we relabel "O" samples based on their distance to the exemplars of each class. Similarly, we define the \textit{NN relabeling threshold} $\mathcal{T}{h}_{NN}$ as:
\begin{equation}
\small
\begin{split}
    \mathcal{T}{h}_{NN}=\beta \cdot  \min_{(x_i,x_j) \in \mathcal{M}_c \atop c\in \mathcal{C}_{t,old}}\{s(h_{t-1}(x_i),h_{t-1}(x_j))\}
\end{split}
\end{equation}
For each "O" sample $x_i$, we then relabel it with $\mathcal{T}{h}_{NN}$ by: 
\begin{equation}
\small
\begin{split}
    \mathcal{S} =&  \{s(h_{t-1}(x_i), h_{t-1}(x_c))\mid x_c \in \mathcal{M}_c, c \in \mathcal{C}_{t,old}\} \\
    y_i=&\mathop{\arg\max}_c \mathcal{S} ,  \quad
    if \max \mathcal{S} > \mathcal{T}{h}_{NN}
\end{split}
\end{equation}

Since the class discrimination of old entity classes keep declining during incremental learning, the older task needs a lower threshold for relabeling sufficient samples. Therefore, we set $\beta_i=0.98-0.05*(t-i)$ for each old task $i$, where $t$ is the current step.

\subsection{Classifying with NCM Classifier}
To make full use of the learned representations, we adopt the Nearest Class Mean (NCM) classifier used in \cite{rebuffi2017icarl} for classification, which is also widely applied in few-shot learning \cite{snell2017prototypical}. For each sample $x$, the class prediction is calculated by:
\begin{equation}
\setlength\abovedisplayskip{3pt}
\setlength\belowdisplayskip{0pt}
    y^*=\mathop{\arg\max}_{c\in \mathcal{C}_{t}^{all}} s(h_t(x), \mathbf{p}_c)
\end{equation}
where $\mathbf{p}_c$ is the prototype of class $c$ calculated with the exemplars as the same in Eq.\ref{prototype}.

\section{Experiment}\label{experiment}

Previous works \cite{monaikul2021continual, xia-etal-2022-learn,wang-etal-2022-shot} on class-incremental NER conducted experiments on the CoNLL 2003 \cite{sang2003introduction} and OntoNotes 5.0 \cite{weischedel2013ontonotes} datasets. However, due to the limited class number of these datasets, the class number introduced in each step and the total number of incremental steps in these datasets are limited. For instance, there are only four classes in the CoNLL03 dataset, thus only one class is introduced in each step and there are only four incremental tasks to repeat. In more realistic situations, multiple classes can be introduced in each step (e.g., a set of product types) and there can be a larger number of incremental steps.

In this work, we provide a \textbf{more realistic and challenging benchmark} for class-incremental NER based on the Few-NERD dataset\footnote{https://ningding97.github.io/fewnerd/} \cite{ding-etal-2021-nerd}, which contains 66 fine-grained entity types. Following the experimental settings of previous works \cite{rebuffi2017icarl, wu2019large,pourkeshavarzi2021looking,madaan2021representational}, we randomly split the 66 classes in Few-NERD into 11 tasks, corresponding to 11 steps, each of which contains 6 entity classes and an "O" class. The training set and development set of each task $\mathcal{T}_t$ contains sentences only labeled with classes of the current task. The test set contains sentences labeled with all learnt classes in task $\{0\dots t\}$. The statistics and class information of each task order can be found in Appendix \ref{dataset_detail}.

\begin{table*}[ht]
\setlength{\tabcolsep}{5pt}
\centering
\small
\begin{tabular}{l|c|c|c|c|c|c|c|c|c|c|c}
\toprule
\multirow{2}*{\bfseries Methods} & \multirow{2}*{\textbf{Step 0}}& \multirow{2}*{\textbf{Step 1}}& \multirow{2}*{\textbf{Step 2}}& \multirow{2}*{\textbf{Step 3}}& \multirow{2}*{\textbf{Step 4}}& \multirow{2}*{\textbf{Step 5}}& \multirow{2}*{\textbf{Step 6}}& \multirow{2}*{\textbf{Step 7}}& \multirow{2}*{\textbf{Step 8}}& \multirow{2}*{\textbf{Step 9}}&
\multirow{2}*{\textbf{Step 10}} \\ 
&&&&&&&&&&\\
\midrule
\bfseries Full Data &
 75.45 & 72.62 & 71.72 & 69.39 & 68.92 & 68.59 & 67.55 & 66.92 & 66.50 & 66.83 & 66.33\\
\midrule

\bfseries LwF & 75.56 & 56.98 & 48.11 & 40.08 & 34.30 & 33.40 & 29.37 & 31.63 & 27.30 & 30.14 & 24.98 \\

\bfseries SCR & 75.14 & 57.39 & 48.73 & 45.47 & 42.76 & 40.94 & 37.75 & 37.49 & 33.59 & 34.51 & 29.54 \\

\bfseries iCaRL & 74.89 & 55.76 & 51.47 & 46.72 & 44.98 & 43.85 & 42.63 & 41.91 & 40.54 & 43.33 & 42.27 \\


\bfseries Con. NER & 75.62 & 55.29 & 42.65 & 35.92 & 32.55 & 30.55 & 26.20 & 27.90 & 25.37 & 28.23 & 25.17 \\
\bfseries Con. NER* & 75.63 & 59.89 & 49.82 & 42.23 & 36.02 & 36.44 & 33.92 & 32.15 & 31.09 & 31.68 & 28.05 \\

\midrule

\bfseries Ours (NN) & 75.73 & \bfseries 65.42 & 62.17 & 56.98 & 55.55 & 52.79 & 51.10 & 49.85 & 47.15 & 49.40 & 47.59 \\
\bfseries Ours (Proto) & 75.73 & 64.98 & \bfseries 62.19 & \bfseries 57.08 & \bfseries 55.56 &\bfseries 54.47 & \bfseries 52.90 & \bfseries 52.16 & \bfseries 51.05 & \bfseries 52.73 & \bfseries 51.16 \\

\bottomrule

\end{tabular}
\caption{Main results of the proposed method and baselines on Few-NERD dataset. For each model, we repeat incremental learning experiments on three different task orders and report the averages of the micro-f1 scores. Detailed results on each task order can be found in Appendix \ref{nerd_detail_result}.}
\label{tab:main}
\vspace{-0.1cm}
\end{table*}

\subsection{Experimental Settings}

The main experiments in this work are conducted on the Few-NERD datasets. Specifically, for each model, we repeat incremental experiments on three different task orders and report the averages of the micro-f1 score. To further illustrate the proposed method on different datasets, we also conduct experiments on the OntoNotes 5.0 dataset (by splitting 18 classes into 6 tasks) in the same way.

We compare our method with 7 comparable baselines. \textbf{Full Data} denotes Bert-tagger \cite{devlin-etal-2019-bert} trained with datasets annotated with both old and new classes, which can be regarded as an upper bound. \textbf{LwF} \cite{li2017learning} is a regularization-based incremental learning method. \textbf{iCaRL} \cite{rebuffi2017icarl} is a typical rehearsal-based representation learning method. \textbf{SCR} \cite{mai2021supervised} is also an effective rehearsal-based contrastive learning method with an NCM classifier. \textbf{Con. NER} or \textbf{Continual NER} \cite{monaikul2021continual} is the previous SOTA method on class-incremental NER. \textbf{Con. NER*} is Continual NER trained with exemplars and tested with NCM classifier. For our method, \textbf{Ours (NN)} and \textbf{Ours (Proto)} denote our method using NN-based and prototype-based strategies, respectively.

The implementation details of baselines and our method, the dataset details, and the detailed macro-f1 and micro-f1 results of different task orders can be found in Appendix \ref{impl_details}, \ref{nerd_detail_result}, \ref{onto_detailed_result} and \ref{dataset_detail}.

\subsection{Main Results}
Table \ref{tab:main} show the results of the proposed method and baselines on the Few-NERD dataset. From the results, we can observe that: (1) The results of \textbf{Full Data}, which leverages all class annotations for training, is relatively consistent. 
(2) Although \textbf{Continual NER} has shown good performance on CoNLL03 or OntoNotes 5.0 datasets, its performance is limited on this more challenging benchmark, when encountering multiple classes and more incremental steps. 
(3) The proposed method shows up to 10.62\% improvement over baselines, and consistently exceeded the baselines by about 10\% even in the later steps, verifying the advantages of the learnt representations. (4) The prototype-based relabeling strategy is more stable than the NN-based strategy especially in the later steps.
A possible reason is that using the mean vector of exemplars for relabeling is more reliable than using each of the exemplars.

We also conduct experiments on the OntoNotes dataset to further illustrate our method.  As shown in Table.\ref{tab:onto}, the results of all methods improve on the less challenging setting, yet the proposed method still significantly outperforms all the baselines.

\begin{table} \setlength{\tabcolsep}{3pt}
\centering
\footnotesize
\scalebox{0.95}{
\begin{tabular}{l|c|c|c|c|c|c}
\toprule
\multirow{2}*{\bfseries Methods} & \multirow{2}*{\textbf{Step0}}& \multirow{2}*{\textbf{Step1}}& \multirow{2}*{\textbf{Step2}}& \multirow{2}*{\textbf{Step3}}& \multirow{2}*{\textbf{Step4}}& \multirow{2}*{\textbf{Step5}} \\ 
&&&&&&\\
\midrule
\bfseries Full Data &
 92.42 & 90.93 & 89.71 & 89.41 & 88.30 & 87.40\\
\midrule

\bfseries LwF & 92.15 & 82.79 & 74.60 & 65.96 & 57.90 & 56.37 \\

\bfseries SCR & 92.04 & 84.80 & 80.53 & 78.67 & 76.48 & 76.05 \\

\bfseries iCaRL & \bfseries 92.49 & 84.89 & 79.65 & 80.03 & 78.76 & 77.14 \\


\bfseries Con. NER & 92.02 & 79.90 & 69.89 & 69.48 & 64.47 & 60.49 \\
\bfseries Con. NER* & 92.09 & 80.23 & 72.23 & 70.88 & 66.65 & 61.48 \\

\midrule

\bfseries Ours (NN) & 92.39 & \bfseries 87.72 & 86.60 & \bfseries 86.83 & 85.64 & 82.98 \\
\bfseries Ours (Proto) & 92.39 & 87.41 & \bfseries 86.87 & 86.42 & \bfseries 85.63 & \bfseries 83.18\\

\bottomrule

\end{tabular}}
\caption{Main results of the proposed method and baselines on OntoNotes 5.0 dataset. We report the averages of the micro-f1 scores of three task orders.}
\label{tab:onto}
\vspace{-0.2cm}
\end{table}

\begin{figure}[ht]
\centering
\includegraphics[width=0.85\linewidth]{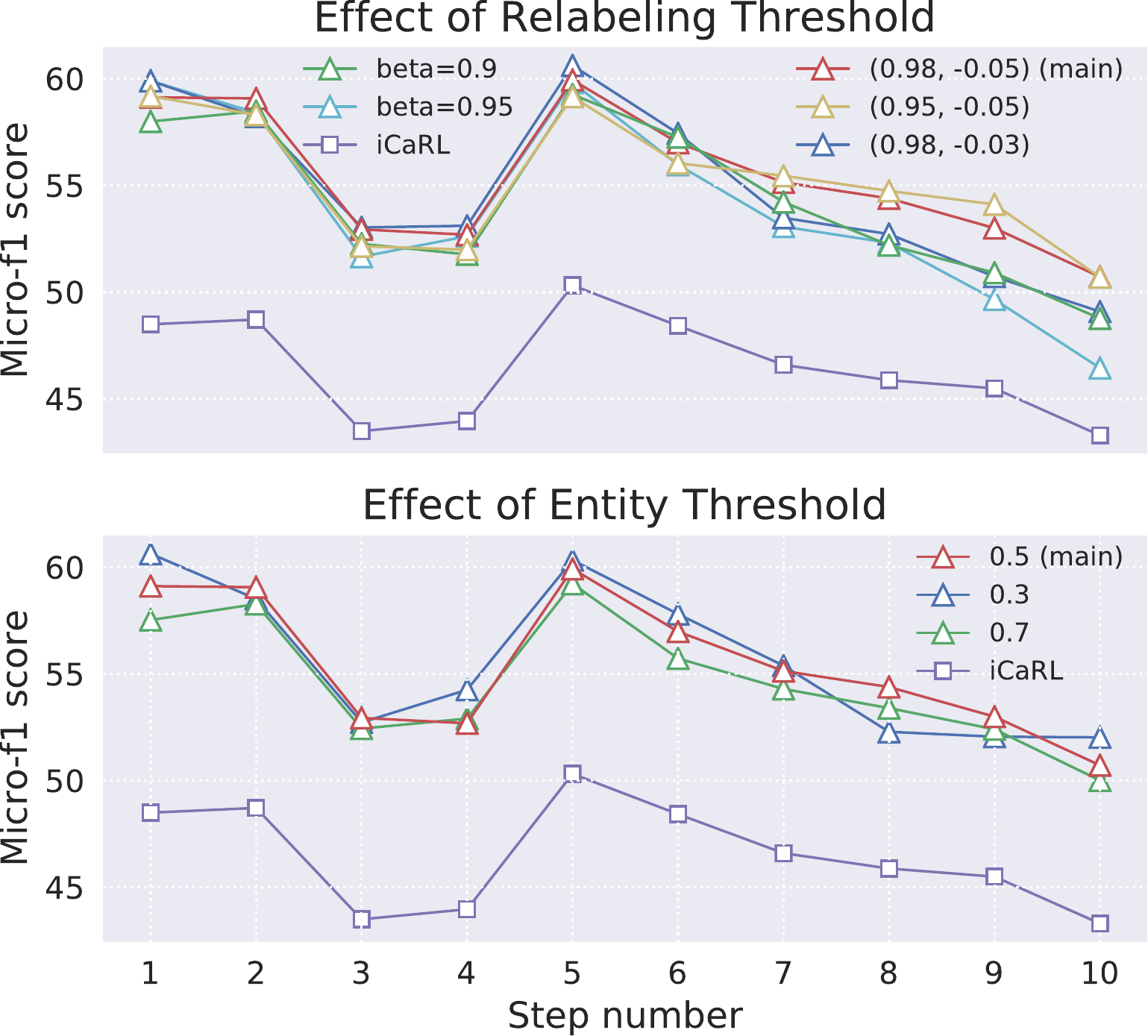}

    \caption{ Effect of different relabeling threshold $\mathcal{T}_{proto}$ and entity threshold $\mathcal{T}_{entity}$. All results are based on the prototype-based method on Few-NERD task order 1.}
    \vspace{-0.2cm}
    \label{threshold}
\end{figure}

\subsection{Ablation Studies}

\begin{figure}
\centering
\includegraphics[width=0.85\linewidth]{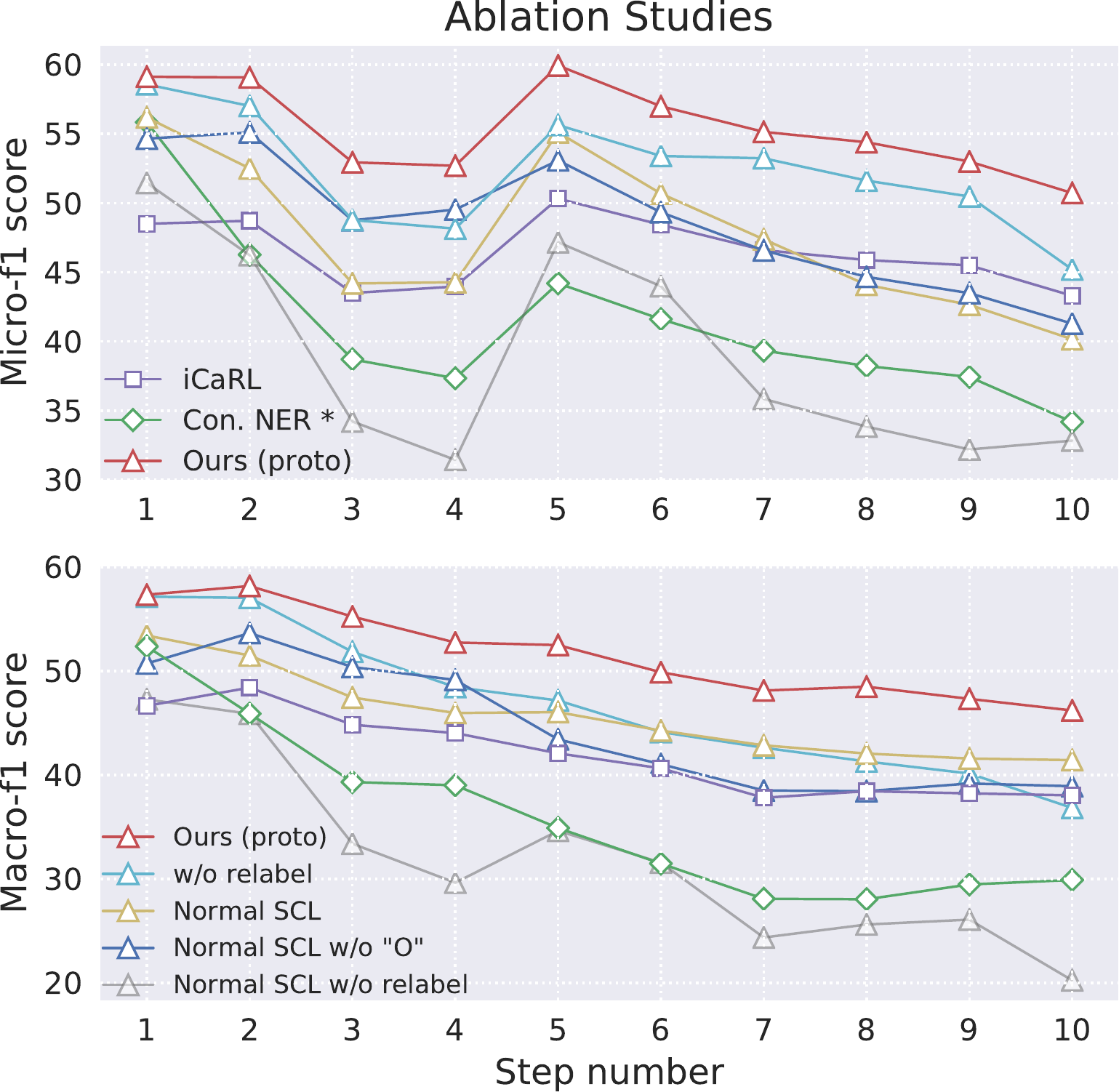}

    \caption{Ablation studies on each component of the proposed method on Few-NERD task order $1$. The upper and the bottom figure shows the micro-f1 and macro-f1 score, respectively. }
    \vspace{-0.2cm}
    \label{ablation}
\end{figure}

To further illustrate the effect of each component on our method, we carry out ablation studies on Few-NERD task order $1$ and show the micro-f1 and macro-f1 results in Figure \ref{ablation}. Here, \textit{Normal SCL} means applying the normal SupCon Loss on both entity classes and "O" without the entity-aware contrastive learning. Similarly, \textit{Normal SCL w/o "O"} means applying the normal SupCon Loss only on entity classes. \textit{Normal SCL w/o relabeling} means applying the normal SupCon Loss without relabel (not using any of our methods). (Both \textit{Normal SCL} and \textit{Normal SCL w/o "O"} adopt prototype-based relabeling) \textit{w/o relabel} denotes using the entity-aware contrastive learning without relabeling.

From the result, we can see that:
(1) Both the relabeling strategy and entity-aware contrastive learning contributes to high performance. (2) The performance of normal SCL without the entity-aware contrastive learning and the relabeling strategy is even worse than iCaRL, indicating that inappropriately learning "O" representations can harm performance. (3) Comparing the micro-f1 and macro-f1 results, we find that the relabeling strategy contributes less to the micro-f1 results. As the micro-f1 results are dominated by head classes with a larger amount of data, we deduce that entity-aware contrastive learning is more useful for head classes (which also appears more in "O"). Also, as the relabeling strategy is based on the distance between representations, the results indicate its effectiveness for both head classes and long-tailed classes.

\subsection{Effect of Threshold Selection}\label{threshold_effect}

Fig.\ref{threshold} shows the results of different hyperparameter choices for threshold calculation. The upper figure refers to the relabeling threshold $\mathcal{T}h_{proto}$, which we set $\beta_i=0.98-0.05*(t-i)$ for each task $t$ in step $i$. In this experiment, we tried different strategies for setting the threshold (\textit{bata=0.9} means $\beta=0.9$, \textit{(0.95,-0.05)} means $\beta_i=0.95-0.05*(t-i)$). We find that the performance is relatively stable w.r.t different choices, and a lower threshold seems more helpful.
\footnote{We further test the relabeling accuracy in Appendix \ref{relabel_stat}.}

In the bottom figure, we also tested for different $\mathcal{T}_{entity}$ choices, which we simply set as the median (0.5) of class similarities. As seen, the performance is also robust to different choices.

\begin{figure}
    \centering
    
\centering
\subfloat[The proposed method]{\includegraphics[width=0.5\linewidth]{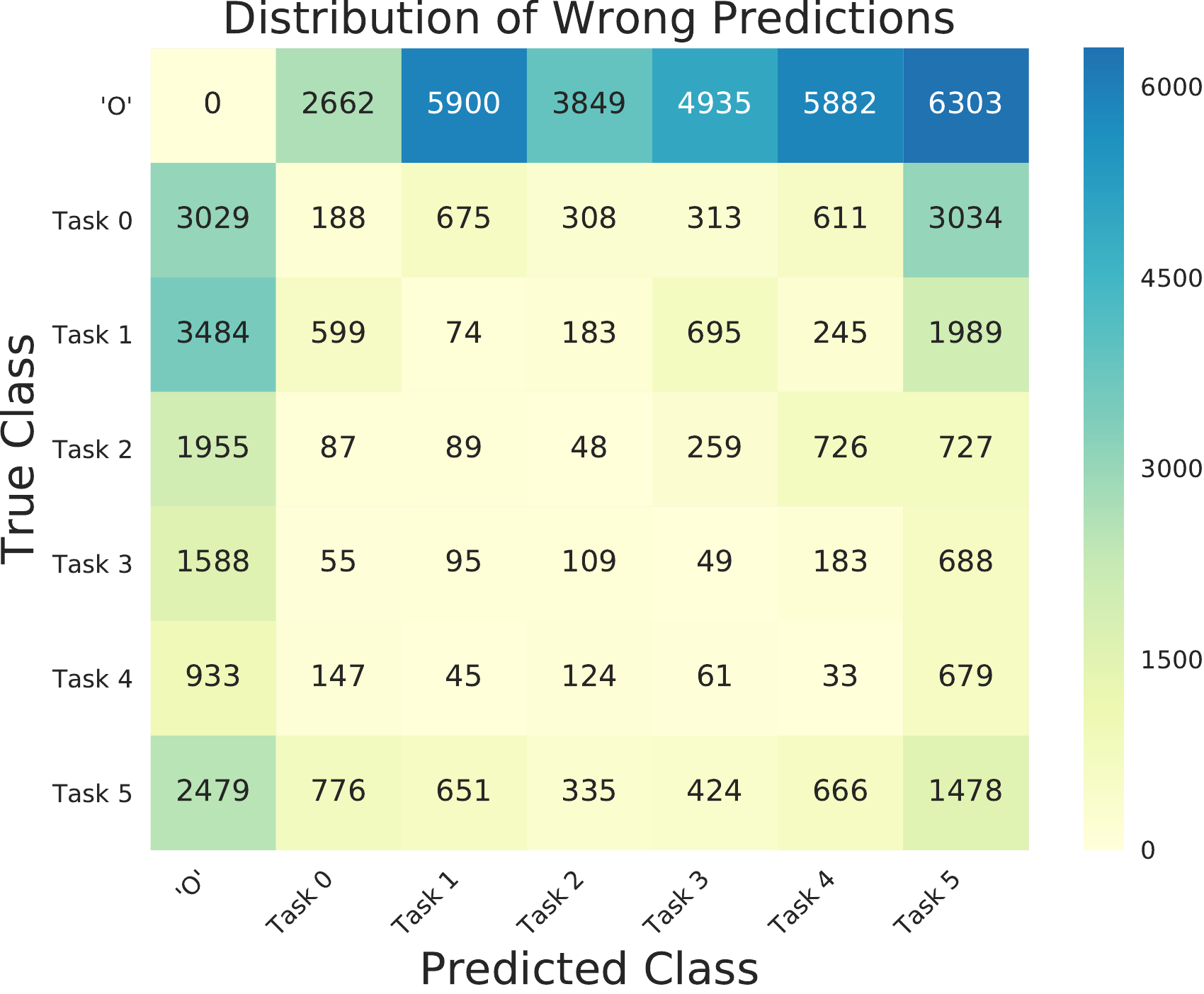}}
\subfloat[Full Data]{\includegraphics[width=0.5\linewidth]{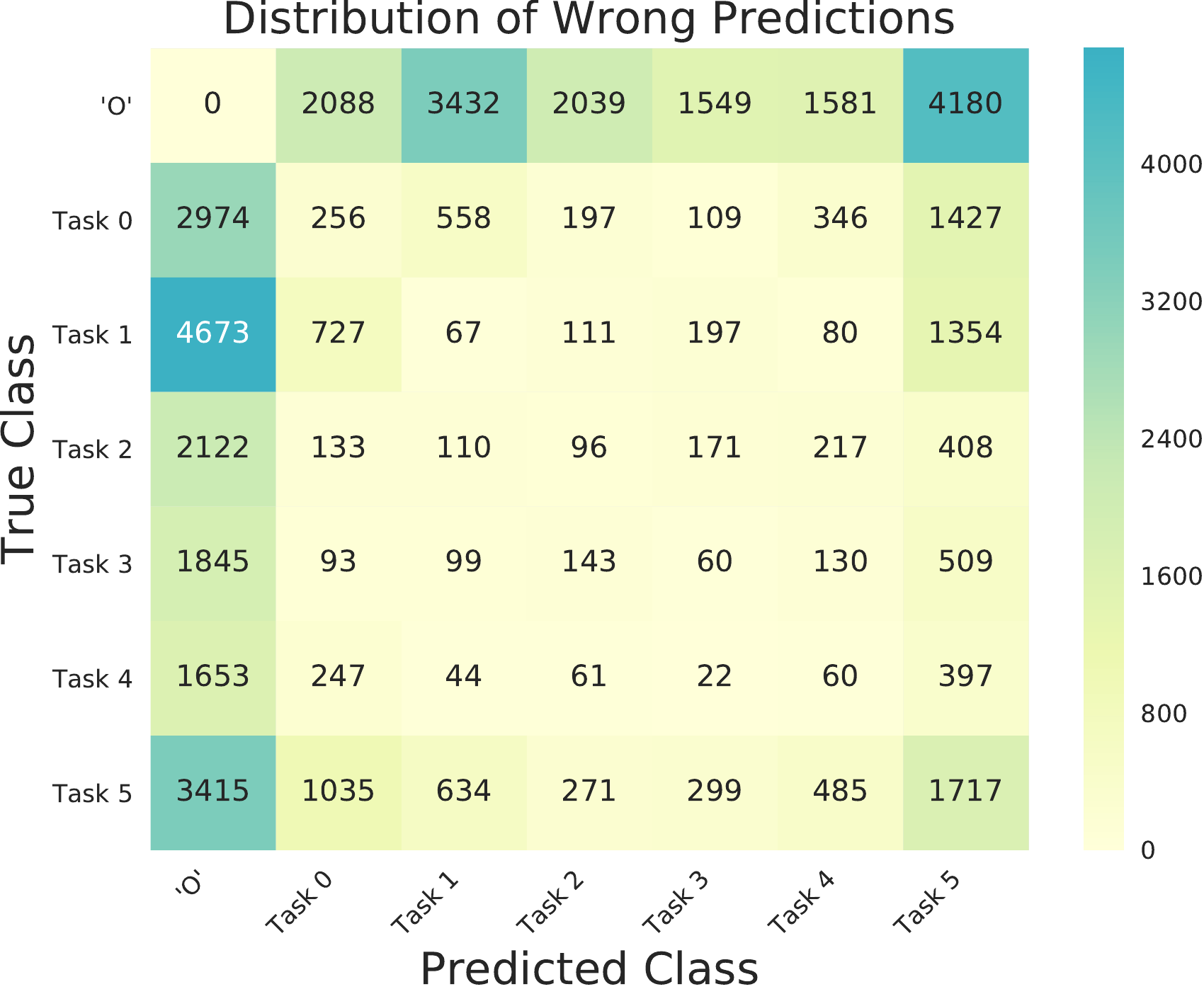}}
    \caption{Distributions of prediction errors of the proposed method and the \textbf{Full Data} baseline in step 6. Compared to Fig.\ref{wrongpred}, the confusion between "O" and entities is largely mitigated, even comparable to \textbf{Full Data}.}
    \label{wrongpred2}
    \vspace{-0.2cm}
\end{figure}

\subsection{Mitigating the Unlabeled Entity Problem}
\begin{figure}
\centering
\includegraphics[width=0.7\linewidth]{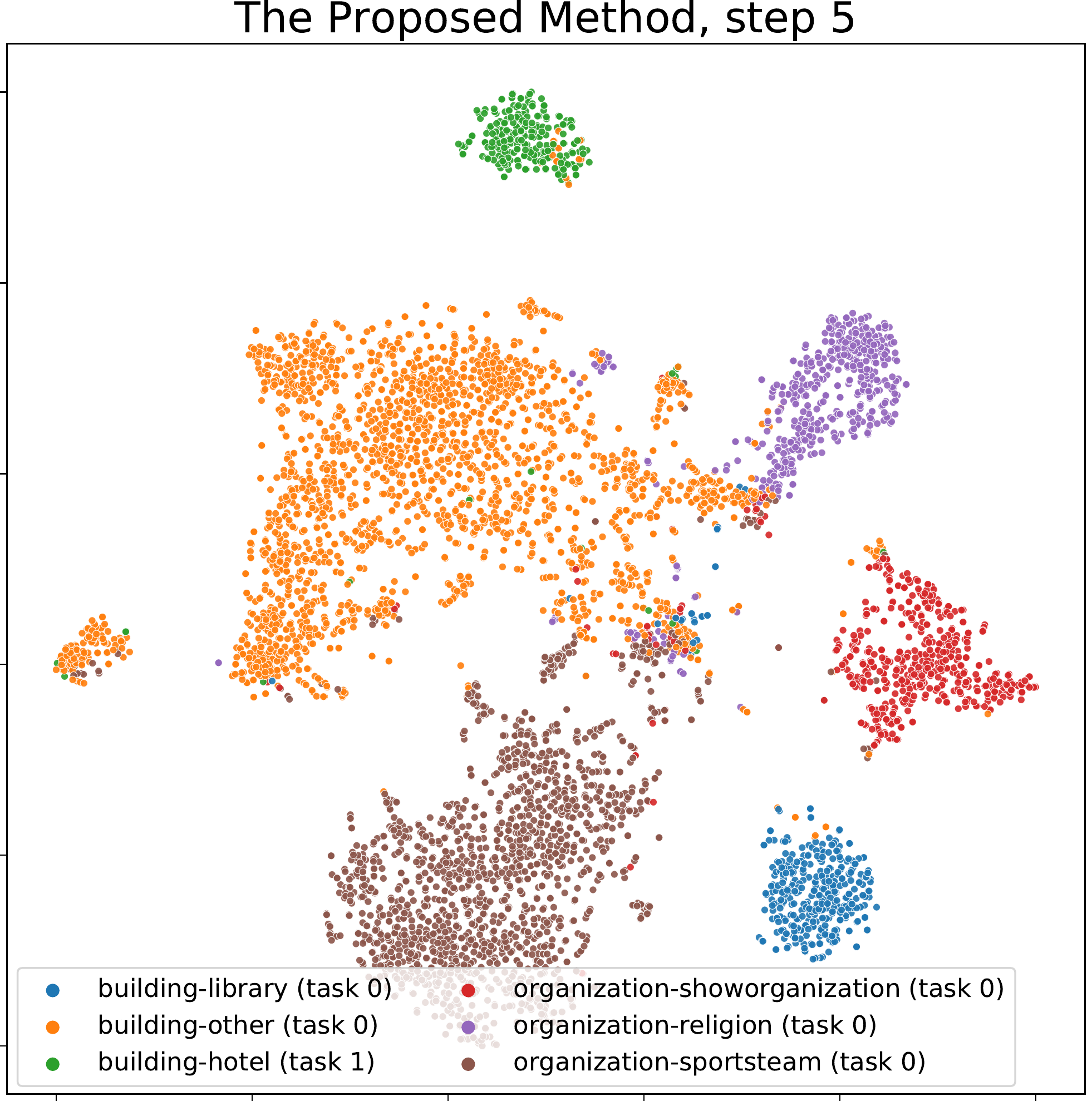}

    \caption{Visualization of the representation of the proposed method in step 5, as a comparison to Fig.\ref{tsne} (b). The proposed method learns much more discriminative representations for the old classes.}
    \vspace{-0.2cm}
    \label{tsne2}
\end{figure}

To demonstrate the effectiveness of the proposed method on mitigating the Unlabeled Entity Problem, we conduct the same experiments as in Section 3. Comparing Fig.\ref{wrongpred2} to Fig.\ref{wrongpred}, we can see that the proposed method largely reduce the confusion between "O" and entities, contributing to much fewer error predictions. Comparing Fig.\ref{tsne2} to Fig.\ref{tsne} (b), we find that the proposed method learns discriminative representations for the old classes despite the impact of incremental learning.

\section{Related Works}
\subsection{Class-incremental Learning}
There are two main research lines of class-incremental learning: (1) Rehearsal-based methods are the most popular and effective methods, which keeps a set of exemplars from the old classes. Typical researches include regularization-based methods that reduces the impact of new classes on old classes \cite{chaudhry2018efficient,riemer2018learning}; methods that aim to alleviate the biased prediction problem in incremental learning \cite{zhao2020maintaining, SaihuiHou2019LearningAU}; methods that replay with generative exemplars \cite{kamra2017deep, ostapenko2019learning,JasonRamapuram2020LifelongGM}. (2) Regularization-based methods aim to regularize the model learning without maintaining any memory. Typical methods include knowledge distillation-based methods \cite{zhang2020class,SaihuiHou2019LearningAU} and gradient-based methods that regularize the model parameters \cite{kirkpatrick2017overcoming, schwarz2018progress, aljundi2018memory}.
These methods, when directly applied to incremental-NER, do not consider the Unlabeled Entity Problem, thus show limited performance. Nonetheless, these methods are essential references for us to improve class-incremental NER.

\subsection{Class-incremental Learning for NER}
Previous works have explored the class-incremental problems in NER \cite{monaikul2021continual,wang-etal-2022-shot,xia-etal-2022-learn}. These methods generally care about maintaining old knowledge. \citet{monaikul2021continual} propose a knowledge distillation-based method for learning old classes in "O". \citet{wang-etal-2022-shot} and \citet{ xia-etal-2022-learn} propose method to generate synthetic samples for old classes. Among these studies, we are the first to comprehensively investigate the Unlabeled Entity Problem and propose solutions that benefits both the old classes and new classes. We also provide a more realistic benchmark.

\subsection{Learning "O" for NER}
Many previous works have also explored "learning 'O'" in NER \cite{tong-etal-2021-learning,li2021empirical,li-etal-2022-rethinking,monaikul2021continual,wang-etal-2022-shot,ma-etal-2022-template}. There are three typical lines of work: (1) \citet{tong-etal-2021-learning} solves the “O” problem for few-shot NER. It proposes a multi-step undefined-class detection approach to explicitly classify potential entity clusters in “O”, which is similar to our core idea. Different from \cite{tong-etal-2021-learning}, we integrate the clustering and detection of potential entity clusters implicitly into representation learning, through a novel design for anchor and positive selection in contrastive learning. To our best knowledge, we are the first to explore the “O” problem in NER with representation learning. 
(2) There also exist other works that study the unlabeled entity problem \cite{li2021empirical,li-etal-2022-rethinking} in NER. These works focus more on avoiding false-negative samples during training and are not specifically designed for distinguishing potential entity classes. 
(3) The ‘O’ problem is also considered by previous works in class-incremental NER \cite{monaikul2021continual,wang-etal-2022-shot}, yet they mainly focus on distilling old knowledge from “O”. Our work provides new insight on the “O” problem (or unlabeled entity problem) by comprehensively considers the old classes and new classes, with detailed experimental results.

\section{Conclusion}
In this work, we first conduct an empirical study to demonstrate the significance of the Unlabeld Entity Problem in class-incremental NER. Based on our observations, we propose a novel and effective representation learning method for learning discriminative representations for "O" and unlabeled entities. To better evaluate class-incremental NER, we introduce a more realistic and challenging benchmark. Intensive experiments demonstrate the effectiveness and show the superior of the proposed method over the baselines.

\section{Limitations}
The limitations of this work are:
(1) In this work, we expect to consider more realistic and more applicable settings for class-incremental NER. Therefore, we consider the Unlabeled Entity Problem and provide a more realistic benchmark based on 66 fine-grained entity types. However, there remain some more serious situations unsolved in this work. First, the entity classes in each step might not be disjoint. For example, a new entity type "Director" might be included in an old entity type "Person". This problem is referred to as the coarse-to-fine problem existing in emerging types of NER. Second, the amount of data or labeled data introduced in each step can also be limited, referring to the few-shot class-incremental problem. Therefore, the proposed method can be further improved to solve these problems. Third, the current version of the proposed method cannot handle the nested NER or contiguous NER problems. In the current version, we simply followed typical works in NER and adopted the sequence labeling scheme to model the NER task, which is not suitable for more complicated NER tasks. Nonetheless, as the proposed representation learning and re-labeling methods are agnostic to the formation of representations, we believe our method can also be adapted to a span-level version, which might be future works. (2) The proposed method is a rehearsal-based method that requires keeping exemplar sets for each class. Although the number of exemplars for each class is really small, we believe there can be more data-efficient solutions that totally avoid the need of memorizing data and also achieve good results.
(3) The proposed method includes several hyper-parameters such as the entity threshold $\mathcal{T}_{entity}$, relabeling threshold $\mathcal{T}h_{NN}$ and $\mathcal{T}h_{proto}$. Although we have shown that the choice of thresholds is relatively robust (Sec.\ref{threshold_effect}), it still requires efforts to explore the most suitable thresholds when applied to other datasets or situations. There can be further work to improve this problem by formulating an automatic threshold searching strategy.

\section*{Acknowledgements}
The authors wish to thank the anonymous reviewers for their helpful comments. This work was partially funded by the National Natural Science Foundation of China (No.62076069,62206057,61976056), Shanghai Rising-Star Program (23QA1400200), and Natural Science Foundation of Shanghai (23ZR1403500).

\bibliography{anthology,custom}
\bibliographystyle{acl_natbib}

\appendix

\newpage 

\section{Appendix}
\label{sec:appendix}

\subsection{Implementation Details}\label{impl_details}
We implemented the proposed method and all baselines based on the \textit{bert-base-cased} pretrained model using the implementation of huggingface transformers \footnote{https://github.com/huggingface/transformers}. For our method, we implement the SupCon loss based on the implementation in the \textit{SupContrast} library\footnote{https://github.com/HobbitLong/SupContrast}. For LwF and iCaRL, we follow the implementations of \cite{masana2020class}~\footnote{https://github.com/mmasana/FACIL}. For SCR, we follow the implementation of the \textit{online-continual-learning
} library\footnote{https://github.com/RaptorMai/online-continual-learning}. There is no public source of Continual NER, so we implement based on the paper \cite{monaikul2021continual} and report the results of our implementation.
At each step, we trained the model for 16 epochs and selected the best model on the dev set. For all methods, we use a learning rate of 5e-5, batch size of 16 and the max sequence length of 128. For our method, we start entity-aware contrastive learning for "O" with $L_{SCL,O}$ at the 10-th epoch and train it for 6 epochs at each step. We conducted all experiments on on NVIDIA GeForce RTX 3090.

\paragraph{Construction of the exemplar set} For all rehearsal-based method, we keep $5$ exemplars for each class, each of which consist of one entity word and its context. The exemplar words of each class are selected by picking the most high-frequency words of each class in the dataset. For each exemplar word, we randomly pick one sentence that contains this word as its context. We use the same exemplar set for all methods.

\subsection{Performance on Old and New Classes}
\begin{figure}
    \centering
    
\centering
\includegraphics[width=1.0\linewidth]{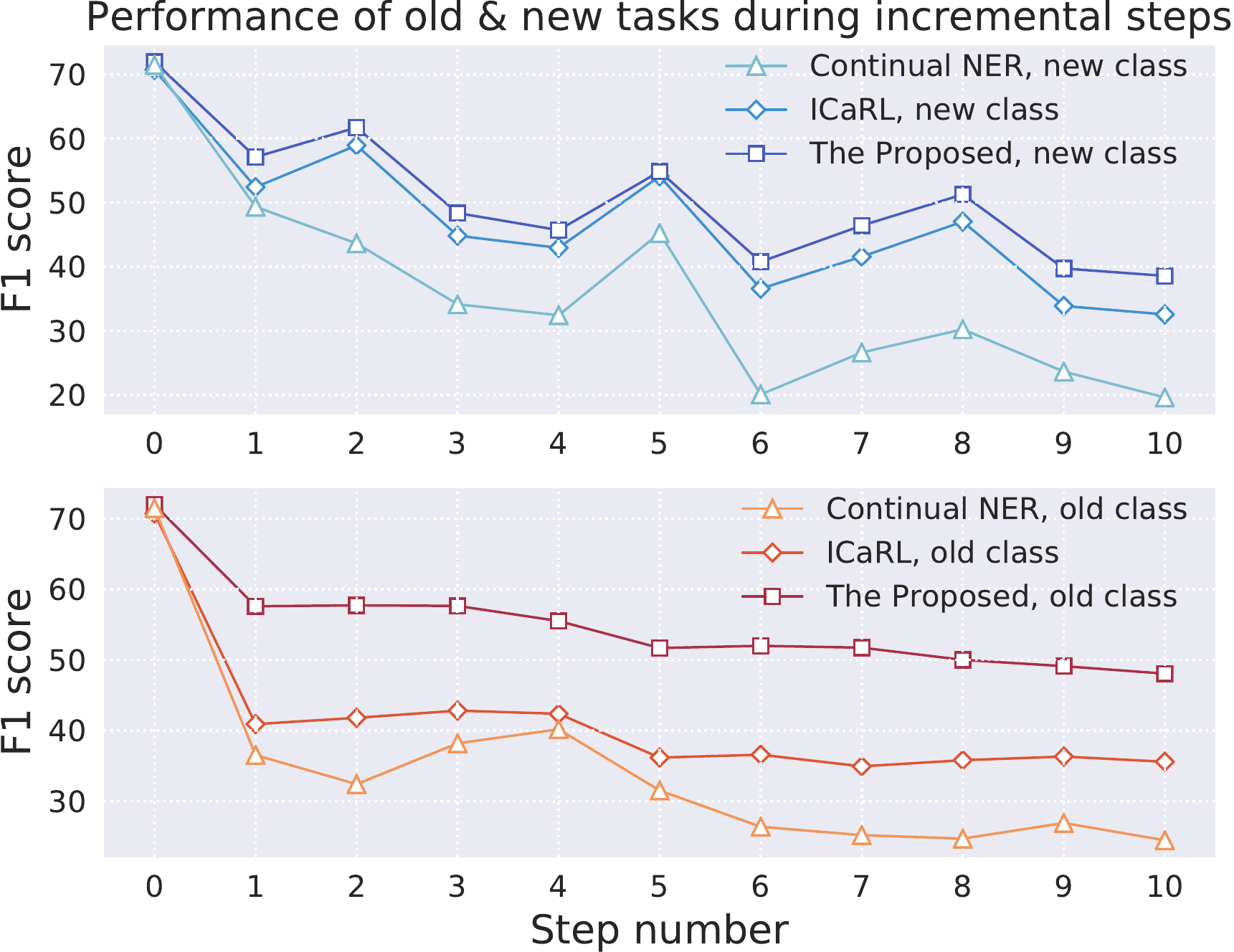}

    \caption{The performance change on old classes and new classes during incremental learning. }
    \vspace{-0.1cm}
    \label{performance}
\end{figure}

In figure \ref{performance}, we show the performance change of different methods on old classes and new classes. As seen, the proposed method can maintain the performance of old classes in a higher degree, which mainly attributes to the relabeling strategy. Meanwhile, the entity-aware contrastive learning method also helps to keep the discrimination of old classes in "O". Also, the proposed method is more effective on learning the new classes than baseline methods, with a highest improvement of $6.01\%$ in the last step. These results indicate the effectiveness of entity-aware contrastive learning, which helps learn fine-grained and entity-aware representations for "O", preventing the potential classes from confusing with "O" and other similar classes.

\begin{table}[ht]
\centering
\begin{tabular}{l|c|c|c}
\toprule
\bfseries Steps  & \bfseries Precision & \bfseries Recall & \bfseries Micro-f1  \\
\midrule

\multicolumn{4}{c}{\textit{Prototype-based relabeling}} \\ 
\midrule

\bfseries Step 1 & 56.61 & 99.04 & 72.04  \\

\bfseries Step 4 & 62.24 & 84.29 & 71.61 \\

\bfseries Step 7 &  74.92 & 70.82 & 72.81 \\

\midrule

\multicolumn{4}{c}{\textit{Prototype-based relabeling ($\beta=0.9$)}} \\ 
\midrule

\bfseries Step 1 & 52.52 & 99.16 & 68.67  \\

\bfseries Step 4 & 61.61 & 73.72 & 67.12 \\

\bfseries Step 7 &  79.40 & 67.63 & 73.05 \\

\midrule

\multicolumn{4}{c}{\textit{NN-based relabeling}} \\ 
\midrule

\bfseries Step 1 & 60.08 &  98.78 & 74.71  \\
\bfseries Step 4 & 64.81 & 81.32 &  72.14 \\
\bfseries Step 7 &  74.56 & 76.55 & 75.54 \\

\bottomrule

\end{tabular}
\caption{Relabeling statistics of different strategies. }
\label{relabel}
\vspace{-0.3cm}
\end{table}

\subsection{Relabeling Statistics}\label{relabel_stat}
We examine the token-level micro-f1 scores of different relabeling strategies based on the gold labeled data of each step on Few-NERD task order 1. The results are shown in Table \ref{relabel}. We find that: (1) The proposed relabeling strategies can achieve acceptable relabeling accuracy, which greatly helps for retaining the knowledge of old classes and improving representation learning for potential classes. (2) Using a fixed $\beta$ leads to higher recall and lower precision in earlier steps, as well as lower recall in later steps. This might because the convergence degree of old classes decrease in later step, thus a fixed threshold will relabel limited number of old class samples. (3) Compared to prototype-based method, the NN-based method has slightly lower recall and higher precision in earlier steps, which might correspond to its slightly higher performance in earlier steps on Few-NERD task order 1 (Table \ref{nerd_result1}).

\subsection{Detailed Results on Few-NERD}\label{nerd_detail_result}
The detailed results on the Few-NERD datasets are shown in Table \ref{nerd_result1} (task order 1), Table \ref{nerd_result2} (task order 2), Table \ref{nerd_result3} (task order 3). In each table, the numbers in black denote the micro-f1 scores and the numbers in green denote the macro-f1 scores. The proposed method surpass all baseline methods in all task orders.

\subsection{Detailed Results on OntoNotes 5.0}
\label{onto_detailed_result}
We also conduct experiments on OntoNotes 5.0\footnote{https://catalog.ldc.upenn.edu/license/ldc-non-members-agreement.pdf} by randomly splitting the 18 entity classes into 6 tasks, each of which contains 3 entity classes and a "O" class. The detailed results on the OntoNotes datasets are shown in Table \ref{onto_result1} (task order 1), Table \ref{onto_result2} (task order 2), Table \ref{onto_result3} (task order 3). In each table, the numbers in black denote the micro-f1 scores and the numbers in green denote the macro-f1 scores. The proposed method surpass all baseline methods in all task orders.

\subsection{Dataset Details}\label{dataset_detail}
The dataset details of Few-NERD are shown in Table \ref{nerd_detail1} (task order 1), Table \ref{nerd_detail2} (task order 2), Table \ref{nerd_detail3} (task order 3). The dataset details of OntoNotes 5.0 are shown in Table \ref{onto_detail1} (task order 1), Table \ref{onto_detail2} (task order 2), Table \ref{onto_detail3} (task order 3). 

\begin{table*}[ht] 
\centering
\small
\begin{tabular}{l|c|c|c|c|c|c|c|c|c|c|c}
\toprule

\multirow{2}*{\bfseries Methods} & \multirow{2}*{\textbf{Step 0}}& \multirow{2}*{\textbf{Step 1}}& \multirow{2}*{\textbf{Step 2}}& \multirow{2}*{\textbf{Step 3}}& \multirow{2}*{\textbf{Step 4}}& \multirow{2}*{\textbf{Step 5}}& \multirow{2}*{\textbf{Step 6}}& \multirow{2}*{\textbf{Step 7}}& \multirow{2}*{\textbf{Step 8}}& \multirow{2}*{\textbf{Step 9}}&
\multirow{2}*{\textbf{Step 10}} \\ 
&&&&&&&&&&\\
\midrule
\multirow{2}*{\bfseries Full Data}
 &
73.30 & 68.73 & 68.02 & 66.21 & 66.60 & 68.79 & 68.40 & 67.29 & 67.44 & 66.61 & 66.35\\
& \textcolor[RGB]{64,130,109}{71.51} & \textcolor[RGB]{64,130,109}{66.21} & \textcolor[RGB]{64,130,109}{67.14} & \textcolor[RGB]{64,130,109}{65.79} & \textcolor[RGB]{64,130,109}{65.20} & \textcolor[RGB]{64,130,109}{61.92} & \textcolor[RGB]{64,130,109}{61.83} & \textcolor[RGB]{64,130,109}{60.24} & \textcolor[RGB]{64,130,109}{60.83} & \textcolor[RGB]{64,130,109}{59.93} & \textcolor[RGB]{64,130,109}{60.32}
 \\

\midrule

\multirow{2}*{\bfseries LwF} 
& 73.47 & 47.63 & 42.96 & 31.85 & 28.98 & 39.73 & 34.64 & 37.66 & 34.16 & 32.23 & 25.47 \\
& \textcolor[RGB]{64,130,109}{71.25} & \textcolor[RGB]{64,130,109}{42.69} & \textcolor[RGB]{64,130,109}{40.58} & \textcolor[RGB]{64,130,109}{31.19} & \textcolor[RGB]{64,130,109}{27.01} & \textcolor[RGB]{64,130,109}{18.97} & \textcolor[RGB]{64,130,109}{19.57} & \textcolor[RGB]{64,130,109}{17.31} & \textcolor[RGB]{64,130,109}{18.49} & \textcolor[RGB]{64,130,109}{17.23} & \textcolor[RGB]{64,130,109}{15.25}
\\
\midrule
\multirow{2}*{\bfseries SCR} & 73.21 & 50.74 & 51.44 & 40.41 & 41.73 & 49.07 & 45.52 & 42.88 & 40.50 & 35.80 & 30.47 \\
& \textcolor[RGB]{64,130,109}{70.56} & \textcolor[RGB]{64,130,109}{46.38} & \textcolor[RGB]{64,130,109}{49.86} & \textcolor[RGB]{64,130,109}{42.67} & \textcolor[RGB]{64,130,109}{39.75} & \textcolor[RGB]{64,130,109}{37.27} & \textcolor[RGB]{64,130,109}{35.15} & \textcolor[RGB]{64,130,109}{32.98} & \textcolor[RGB]{64,130,109}{29.90} & \textcolor[RGB]{64,130,109}{29.12} & \textcolor[RGB]{64,130,109}{27.86}
\\
\midrule

\multirow{2}*{\bfseries iCaRL} & 72.69 & 48.50 & 48.72 & 43.50 & 43.97 & 50.32 & 48.43 & 46.60 & 45.88 & 45.5 & 43.30 \\
& \textcolor[RGB]{64,130,109}{70.71} & \textcolor[RGB]{64,130,109}{46.66} & \textcolor[RGB]{64,130,109}{48.41} & \textcolor[RGB]{64,130,109}{44.84} & \textcolor[RGB]{64,130,109}{44.04} & \textcolor[RGB]{64,130,109}{42.09} & \textcolor[RGB]{64,130,109}{40.65} & \textcolor[RGB]{64,130,109}{37.82} & \textcolor[RGB]{64,130,109}{38.45} & \textcolor[RGB]{64,130,109}{38.24} & \textcolor[RGB]{64,130,109}{38.05}
\\
\midrule


\multirow{2}*{\bfseries Con.NER}  & 73.42 & 47.02 & 43.09 & 35.86 & 36.47 & 44.79 & 37.49 & 37.08 & 36.43 & 35.24 & 27.04 \\
& \textcolor[RGB]{64,130,109}{71.45} & \textcolor[RGB]{64,130,109}{42.93} & \textcolor[RGB]{64,130,109}{39.87} & \textcolor[RGB]{64,130,109}{34.82} & \textcolor[RGB]{64,130,109}{34.60} & \textcolor[RGB]{64,130,109}{30.73} & \textcolor[RGB]{64,130,109}{24.90} & \textcolor[RGB]{64,130,109}{20.91} & \textcolor[RGB]{64,130,109}{22.20} & \textcolor[RGB]{64,130,109}{22.53} & \textcolor[RGB]{64,130,109}{20.33}
\\
\midrule
\multirow{2}*{\bfseries Con.NER*} & 73.56 & 55.84 & 46.27 & 38.71 & 37.34 & 44.20 & 41.61 & 39.34 & 38.23 & 37.44 & 34.19 \\
& \textcolor[RGB]{64,130,109}{71.90} & \textcolor[RGB]{64,130,109}{52.37} & \textcolor[RGB]{64,130,109}{45.91} & \textcolor[RGB]{64,130,109}{39.33} & \textcolor[RGB]{64,130,109}{39.02} & \textcolor[RGB]{64,130,109}{34.91} & \textcolor[RGB]{64,130,109}{31.49} & \textcolor[RGB]{64,130,109}{28.10} & \textcolor[RGB]{64,130,109}{28.06} & \textcolor[RGB]{64,130,109}{29.48} & \textcolor[RGB]{64,130,109}{29.91}
\\

\midrule

\multirow{2}*{\bfseries Ours (NN)} & \textbf{74.04} & \textbf{59.22} & \textbf{59.08} & 52.18 & \textbf{53.24} & \textbf{60.51} & \textbf{57.81} & \textbf{55.41} & 53.61 & 49.44 & 46.93 \\
& \textbf{\textcolor[RGB]{64,130,109}{71.94}} & \textbf{\textcolor[RGB]{64,130,109}{57.40}} & \textbf{\textcolor[RGB]{64,130,109}{58.36}} & \textcolor[RGB]{64,130,109}{54.64} & \textbf{\textcolor[RGB]{64,130,109}{53.35}} & \textbf{\textcolor[RGB]{64,130,109}{52.87}} & \textbf{\textcolor[RGB]{64,130,109}{50.60}} & \textbf{\textcolor[RGB]{64,130,109}{48.48}} & \textcolor[RGB]{64,130,109}{47.97} & \textcolor[RGB]{64,130,109}{45.30} & \textcolor[RGB]{64,130,109}{44.80}
\\
\midrule
\multirow{2}*{\bfseries Ours (Proto)} & \textbf{74.04} & 59.12 & 59.07 & \textbf{52.94} & 52.69 & 59.93 & 56.99 & 55.14 & \textbf{54.39} & \textbf{53.00} & \textbf{50.72} \\
& \textbf{\textcolor[RGB]{64,130,109}{71.94}} & \textcolor[RGB]{64,130,109}{57.35} & \textcolor[RGB]{64,130,109}{58.18} & \textbf{\textcolor[RGB]{64,130,109}{55.24}} & \textcolor[RGB]{64,130,109}{52.75} & \textcolor[RGB]{64,130,109}{52.50} & \textcolor[RGB]{64,130,109}{49.89} & \textcolor[RGB]{64,130,109}{48.13} & \textbf{\textcolor[RGB]{64,130,109}{48.50}} & \textbf{\textcolor[RGB]{64,130,109}{47.33}} & \textbf{\textcolor[RGB]{64,130,109}{46.21}}
\\

\bottomrule

\end{tabular}
\caption{Detailed results of Few-NERD task order 1. The numbers in black are the micro-f1 scores and the numbers in green are the macro-f1 scores.}
\label{nerd_result1}
\end{table*}

\begin{table*}[ht] 
\centering
\small
\begin{tabular}{l|c|c|c|c|c|c|c|c|c|c|c}
\toprule

\multirow{2}*{\bfseries Methods} & \multirow{2}*{\textbf{Step 0}}& \multirow{2}*{\textbf{Step 1}}& \multirow{2}*{\textbf{Step 2}}& \multirow{2}*{\textbf{Step 3}}& \multirow{2}*{\textbf{Step 4}}& \multirow{2}*{\textbf{Step 5}}& \multirow{2}*{\textbf{Step 6}}& \multirow{2}*{\textbf{Step 7}}& \multirow{2}*{\textbf{Step 8}}& \multirow{2}*{\textbf{Step 9}}&
\multirow{2}*{\textbf{Step 10}} \\ 
&&&&&&&&&&\\
\midrule
\multirow{2}*{\bfseries Full Data}
 &
82.26 & 78.9 & 77.47 & 73.02 & 71.57 & 69.10 & 68.08 & 67.67 & 67.32 & 66.48 & 66.29 \\
& \textcolor[RGB]{64,130,109}{68.47} & \textcolor[RGB]{64,130,109}{68.24} & \textcolor[RGB]{64,130,109}{68.11} & \textcolor[RGB]{64,130,109}{63.38} & \textcolor[RGB]{64,130,109}{63.26} & \textcolor[RGB]{64,130,109}{61.19} & \textcolor[RGB]{64,130,109}{60.72} & \textcolor[RGB]{64,130,109}{61.05} & \textcolor[RGB]{64,130,109}{60.31} & \textcolor[RGB]{64,130,109}{60.12} & \textcolor[RGB]{64,130,109}{60.26}
 \\

\midrule

\multirow{2}*{\bfseries LwF} 
& 82.33 & 67.39 & 55.59 & 49.60 & 37.25 & 30.90 & 28.86 & 32.30 & 29.35 & 27.86 & 23.57
 \\
& \textcolor[RGB]{64,130,109}{68.15} & \textcolor[RGB]{64,130,109}{53.57} & \textcolor[RGB]{64,130,109}{42.70} & \textcolor[RGB]{64,130,109}{34.89} & \textcolor[RGB]{64,130,109}{26.02} & \textcolor[RGB]{64,130,109}{22.06} & \textcolor[RGB]{64,130,109}{18.49} & \textcolor[RGB]{64,130,109}{16.71} & \textcolor[RGB]{64,130,109}{15.63} & \textcolor[RGB]{64,130,109}{14.24} & \textcolor[RGB]{64,130,109}{12.35}
\\
\midrule
\multirow{2}*{\bfseries SCR} & 81.95 & 67.68 & 50.66 & 50.98 & 42.82 & 34.78 & 34.11 & 37.90 & 31.68 & 29.40 & 23.82 \\
& \textcolor[RGB]{64,130,109}{66.04} & \textcolor[RGB]{64,130,109}{53.06} & \textcolor[RGB]{64,130,109}{43.86} & \textcolor[RGB]{64,130,109}{41.92} & \textcolor[RGB]{64,130,109}{31.19} & \textcolor[RGB]{64,130,109}{29.55} & \textcolor[RGB]{64,130,109}{29.48} & \textcolor[RGB]{64,130,109}{30.00} & \textcolor[RGB]{64,130,109}{26.78} & \textcolor[RGB]{64,130,109}{24.95} & \textcolor[RGB]{64,130,109}{22.80}
\\
\midrule

\multirow{2}*{\bfseries iCaRL} & 81.91 & 66.01 & 56.73 & 51.19 & 45.16 & 40.64 & 41.05 & 41.52 & 41.13 & 41.58 & 41.50 \\
& \textcolor[RGB]{64,130,109}{67.28} & \textcolor[RGB]{64,130,109}{52.06} & \textcolor[RGB]{64,130,109}{44.50} & \textcolor[RGB]{64,130,109}{40.92} & \textcolor[RGB]{64,130,109}{39.07} & \textcolor[RGB]{64,130,109}{37.61} & \textcolor[RGB]{64,130,109}{38.34} & \textcolor[RGB]{64,130,109}{37.10} & \textcolor[RGB]{64,130,109}{37.22} & \textcolor[RGB]{64,130,109}{37.64} & \textcolor[RGB]{64,130,109}{38.07}
\\
\midrule


\multirow{2}*{\bfseries Con.NER}  & \textbf{82.44} & 66.55 & 42.94 & 38.07 & 29.65 & 24.31 & 22.07 & 25.91 & 24.59 & 23.62 & 23.35 \\
& \textbf{\textcolor[RGB]{64,130,109}{68.91}} & \textcolor[RGB]{64,130,109}{53.03} & \textcolor[RGB]{64,130,109}{35.11} & \textcolor[RGB]{64,130,109}{31.56} & \textcolor[RGB]{64,130,109}{26.61} & \textcolor[RGB]{64,130,109}{23.20} & \textcolor[RGB]{64,130,109}{19.22} & \textcolor[RGB]{64,130,109}{18.13} & \textcolor[RGB]{64,130,109}{17.80} & \textcolor[RGB]{64,130,109}{15.96} & \textcolor[RGB]{64,130,109}{17.73}
\\
\midrule
\multirow{2}*{\bfseries Con.NER*} & 82.38 & 68.89 & 56.38 & 48.42 & 37.71 & 33.76 & 33.81 & 29.43 & 29.84 & 28.89 & 26.75 \\
& \textcolor[RGB]{64,130,109}{68.88} & \textcolor[RGB]{64,130,109}{55.10} & \textcolor[RGB]{64,130,109}{44.16} & \textcolor[RGB]{64,130,109}{39.21} & \textcolor[RGB]{64,130,109}{35.69} & \textcolor[RGB]{64,130,109}{32.16} & \textcolor[RGB]{64,130,109}{30.21} & \textcolor[RGB]{64,130,109}{25.66} & \textcolor[RGB]{64,130,109}{28.06} & \textcolor[RGB]{64,130,109}{26.39} & \textcolor[RGB]{64,130,109}{26.92}
\\

\midrule

\multirow{2}*{\bfseries Ours (NN)} & 82.32 & \textbf{73.27} & 67.96 & \textbf{62.07} & \textbf{57.49} & 47.27 & 48.95 & 52.33 & 49.75 & 49.44 & 49.75 \\
& \textcolor[RGB]{64,130,109}{68.29} & \textbf{\textcolor[RGB]{64,130,109}{60.79}} & \textcolor[RGB]{64,130,109}{55.97} & \textbf{\textcolor[RGB]{64,130,109}{52.71}} & \textbf{\textcolor[RGB]{64,130,109}{50.28}} & \textcolor[RGB]{64,130,109}{46.88} & \textcolor[RGB]{64,130,109}{47.08} & \textcolor[RGB]{64,130,109}{45.34} & \textcolor[RGB]{64,130,109}{46.58} & \textcolor[RGB]{64,130,109}{44.72} & \textcolor[RGB]{64,130,109}{44.17}
\\
\midrule
\multirow{2}*{\bfseries Ours (Proto)} & 82.32 & 72.97 & \textbf{68.38} & 61.64 & 56.75 & \textbf{51.30} & \textbf{53.33} & \textbf{53.44} & \textbf{52.75} & \textbf{52.20} & \textbf{52.10}  \\
& \textcolor[RGB]{64,130,109}{68.29} & \textcolor[RGB]{64,130,109}{59.84} & \textbf{\textcolor[RGB]{64,130,109}{56.07}} & \textcolor[RGB]{64,130,109}{52.36} & \textcolor[RGB]{64,130,109}{49.74} & \textbf{\textcolor[RGB]{64,130,109}{47.26}} & \textbf{\textcolor[RGB]{64,130,109}{47.73}} & \textbf{\textcolor[RGB]{64,130,109}{47.67}} & \textbf{\textcolor[RGB]{64,130,109}{46.72}} & \textbf{\textcolor[RGB]{64,130,109}{46.50}} & \textbf{\textcolor[RGB]{64,130,109}{46.06}}
\\

\bottomrule

\end{tabular}
\caption{Detailed results of Few-NERD task order 2. The numbers in black are the micro-f1 scores and the numbers in green are the macro-f1 scores.}
\label{nerd_result2}
\end{table*}

\begin{table*}[ht] 
\centering
\small
\begin{tabular}{l|c|c|c|c|c|c|c|c|c|c|c}
\toprule

\multirow{2}*{\bfseries Methods} & \multirow{2}*{\textbf{Step 0}}& \multirow{2}*{\textbf{Step 1}}& \multirow{2}*{\textbf{Step 2}}& \multirow{2}*{\textbf{Step 3}}& \multirow{2}*{\textbf{Step 4}}& \multirow{2}*{\textbf{Step 5}}& \multirow{2}*{\textbf{Step 6}}& \multirow{2}*{\textbf{Step 7}}& \multirow{2}*{\textbf{Step 8}}& \multirow{2}*{\textbf{Step 9}}&
\multirow{2}*{\textbf{Step 10}} \\ 
&&&&&&&&&&\\
\midrule
\multirow{2}*{\bfseries Full Data}
 &
70.79 & 70.24 & 69.68 & 68.94 & 68.58 & 67.87 & 66.16 & 65.81 & 64.74 & 67.40 & 66.35 \\
& \textcolor[RGB]{64,130,109}{72.21} & \textcolor[RGB]{64,130,109}{69.30} & \textcolor[RGB]{64,130,109}{68.62} & \textcolor[RGB]{64,130,109}{63.63} & \textcolor[RGB]{64,130,109}{64.17} & \textcolor[RGB]{64,130,109}{63.84} & \textcolor[RGB]{64,130,109}{63.30} & \textcolor[RGB]{64,130,109}{63.25} & \textcolor[RGB]{64,130,109}{62.18} & \textcolor[RGB]{64,130,109}{61.09} & \textcolor[RGB]{64,130,109}{60.47}
 \\

\midrule

\multirow{2}*{\bfseries LwF} 
& 70.87 & 55.91 & 45.79 & 38.79 & 36.68 & 29.58 & 24.61 & 24.93 & 18.39 & 30.32 & 25.90
 \\
& \textcolor[RGB]{64,130,109}{72.12} & \textcolor[RGB]{64,130,109}{49.67} & \textcolor[RGB]{64,130,109}{36.79} & \textcolor[RGB]{64,130,109}{27.59} & \textcolor[RGB]{64,130,109}{27.72} & \textcolor[RGB]{64,130,109}{24.37} & \textcolor[RGB]{64,130,109}{19.31} & \textcolor[RGB]{64,130,109}{20.93} & \textcolor[RGB]{64,130,109}{18.43} & \textcolor[RGB]{64,130,109}{16.06} & \textcolor[RGB]{64,130,109}{14.51}
\\
\midrule
\multirow{2}*{\bfseries SCR} & 70.25 & 53.74 & 44.09 & 45.02 & 43.72 & 38.98 & 33.62 & 31.70 & 28.60 & 38.33 & 34.32 \\
& \textcolor[RGB]{64,130,109}{70.47} & \textcolor[RGB]{64,130,109}{54.34} & \textcolor[RGB]{64,130,109}{43.22} & \textcolor[RGB]{64,130,109}{35.97} & \textcolor[RGB]{64,130,109}{36.57} & \textcolor[RGB]{64,130,109}{33.32} & \textcolor[RGB]{64,130,109}{31.56} & \textcolor[RGB]{64,130,109}{31.11} & \textcolor[RGB]{64,130,109}{28.49} & \textcolor[RGB]{64,130,109}{27.76} & \textcolor[RGB]{64,130,109}{26.76}
\\
\midrule

\multirow{2}*{\bfseries iCaRL} & 70.07 & 52.78 & 48.98 & 45.48 & 45.81 & 40.57 & 38.42 & 37.60 & 34.60 & 42.92 & 42.00 \\
& \textcolor[RGB]{64,130,109}{70.64} & \textcolor[RGB]{64,130,109}{52.63} & \textcolor[RGB]{64,130,109}{47.93} & \textcolor[RGB]{64,130,109}{40.26} & \textcolor[RGB]{64,130,109}{40.20} & \textcolor[RGB]{64,130,109}{38.33} & \textcolor[RGB]{64,130,109}{38.36} & \textcolor[RGB]{64,130,109}{38.41} & \textcolor[RGB]{64,130,109}{37.74} & \textcolor[RGB]{64,130,109}{37.73} & \textcolor[RGB]{64,130,109}{38.25}
\\
\midrule


\multirow{2}*{\bfseries Con.NER}  & \textbf{70.98} & 52.28 & 41.93 & 33.84 & 31.51 & 22.54 & 19.05 & 20.70 & 15.10 & 25.85 & 25.13 \\
& \textbf{\textcolor[RGB]{64,130,109}{72.82}} & \textcolor[RGB]{64,130,109}{48.64} & \textcolor[RGB]{64,130,109}{36.63} & \textcolor[RGB]{64,130,109}{21.51} & \textcolor[RGB]{64,130,109}{22.83} & \textcolor[RGB]{64,130,109}{20.34} & \textcolor[RGB]{64,130,109}{16.07} & \textcolor[RGB]{64,130,109}{17.23} & \textcolor[RGB]{64,130,109}{15.81} & \textcolor[RGB]{64,130,109}{11.25} & \textcolor[RGB]{64,130,109}{11.51}
\\
\midrule
\multirow{2}*{\bfseries Con.NER*} & 70.95 & 54.95 & 46.81 & 39.57 & 33.03 & 31.36 & 26.34 & 27.68 & 25.22 & 28.71 & 23.20 \\
& \textcolor[RGB]{64,130,109}{72.71} & \textcolor[RGB]{64,130,109}{54.01} & \textcolor[RGB]{64,130,109}{44.32} & \textcolor[RGB]{64,130,109}{32.84} & \textcolor[RGB]{64,130,109}{31.78} & \textcolor[RGB]{64,130,109}{29.89} & \textcolor[RGB]{64,130,109}{27.08} & \textcolor[RGB]{64,130,109}{27.84} & \textcolor[RGB]{64,130,109}{26.25} & \textcolor[RGB]{64,130,109}{19.18} & \textcolor[RGB]{64,130,109}{20.75}
\\

\midrule

\multirow{2}*{\bfseries Ours (NN)} & 70.84 & \textbf{63.77} & \textbf{59.47} & \textbf{56.68} & 55.93 & 50.58 & 46.54 & 41.80 & 38.09 & 49.31 & 46.10 \\
& \textcolor[RGB]{64,130,109}{72.35} & \textbf{\textcolor[RGB]{64,130,109}{63.73}} & \textbf{\textcolor[RGB]{64,130,109}{58.15}} & \textcolor[RGB]{64,130,109}{51.87} & \textcolor[RGB]{64,130,109}{51.64} & \textcolor[RGB]{64,130,109}{50.37} & \textcolor[RGB]{64,130,109}{47.79} & \textcolor[RGB]{64,130,109}{43.92} & \textcolor[RGB]{64,130,109}{43.86} & \textcolor[RGB]{64,130,109}{44.58} & \textcolor[RGB]{64,130,109}{44.62}
\\
\midrule
\multirow{2}*{\bfseries Ours (Proto)} & 70.84 & 62.85 & 59.12 & 56.66 & \textbf{57.23} & \textbf{52.18} & \textbf{48.38} & \textbf{47.91} & \textbf{46.02} & \textbf{52.99} & \textbf{50.68}  \\
& \textcolor[RGB]{64,130,109}{72.35} & \textcolor[RGB]{64,130,109}{62.23} & \textcolor[RGB]{64,130,109}{57.54} & \textbf{\textcolor[RGB]{64,130,109}{52.32}} & \textbf{\textcolor[RGB]{64,130,109}{53.80}} & \textbf{\textcolor[RGB]{64,130,109}{51.79}} & \textbf{\textcolor[RGB]{64,130,109}{49.97}} & \textbf{\textcolor[RGB]{64,130,109}{50.42}} & \textbf{\textcolor[RGB]{64,130,109}{49.21}} & \textbf{\textcolor[RGB]{64,130,109}{49.25}} & \textbf{\textcolor[RGB]{64,130,109}{48.11}}
\\

\bottomrule

\end{tabular}
\caption{Detailed results of Few-NERD task order 3. The numbers in black are the micro-f1 scores and the numbers in green are the macro-f1 scores.}
\label{nerd_result3}
\end{table*}

\begin{table*}[ht] 
\centering
\small
\begin{tabular}{l|c|c|c|c|c|c}
\toprule

\multirow{2}*{\bfseries Methods} & \multirow{2}*{\textbf{Step 0}}& \multirow{2}*{\textbf{Step 1}}& \multirow{2}*{\textbf{Step 2}}& \multirow{2}*{\textbf{Step 3}}& \multirow{2}*{\textbf{Step 4}}& \multirow{2}*{\textbf{Step 5}} \\ 
&&&&&\\
\midrule
\multirow{2}*{\bfseries Full Data}
 &
93.71 & 91.07 & 90.97 & 90.38 & 88.93 & 87.47\\
& \textcolor[RGB]{64,130,109}{84.82} & \textcolor[RGB]{64,130,109}{82.45} & \textcolor[RGB]{64,130,109}{78.11} & \textcolor[RGB]{64,130,109}{78.36} & \textcolor[RGB]{64,130,109}{77.02} & \textcolor[RGB]{64,130,109}{76.97}
 \\

\midrule

\multirow{2}*{\bfseries LwF} 
& 93.28 & 86.63 & 73.58 & 73.60 & 71.70 & 64.24 \\
& \textbf{\textcolor[RGB]{64,130,109}{84.74}} & \textcolor[RGB]{64,130,109}{77.11} & \textcolor[RGB]{64,130,109}{63.58} & \textcolor[RGB]{64,130,109}{61.68} & \textcolor[RGB]{64,130,109}{53.73} & \textcolor[RGB]{64,130,109}{52.86}
\\
\midrule
\multirow{2}*{\bfseries SCR} & 93.36 & 89.28 & 86.10 & 82.81 & 81.98 & 78.47 \\
& \textcolor[RGB]{64,130,109}{81.71} & \textcolor[RGB]{64,130,109}{75.59} & \textcolor[RGB]{64,130,109}{71.50} & \textcolor[RGB]{64,130,109}{69.06} & \textcolor[RGB]{64,130,109}{67.28} & \textcolor[RGB]{64,130,109}{66.75}

\\
\midrule

\multirow{2}*{\bfseries iCaRL} & 93.62 & 87.78 & 78.91 & 79.60 & 76.52 & 75.33 \\
& \textcolor[RGB]{64,130,109}{84.21} & \textcolor[RGB]{64,130,109}{78.09} & \textcolor[RGB]{64,130,109}{65.35} & \textcolor[RGB]{64,130,109}{68.64} & \textcolor[RGB]{64,130,109}{65.08} & \textcolor[RGB]{64,130,109}{65.07}
\\
\midrule


\multirow{2}*{\bfseries Con.NER}  & 93.16 & 83.25 & 70.90 & 71.59 & 60.26 & 63.15 \\
& \textcolor[RGB]{64,130,109}{83.62} & \textcolor[RGB]{64,130,109}{71.99} & \textcolor[RGB]{64,130,109}{59.90} & \textcolor[RGB]{64,130,109}{59.01} & \textcolor[RGB]{64,130,109}{50.13} & \textcolor[RGB]{64,130,109}{48.28}
\\
\midrule
\multirow{2}*{\bfseries Con.NER*} & 93.24 & 83.53 & 73.81 & 72.25 & 64.03 & 62.42 \\
& \textcolor[RGB]{64,130,109}{83.46} & \textcolor[RGB]{64,130,109}{72.51} & \textcolor[RGB]{64,130,109}{60.29} & \textcolor[RGB]{64,130,109}{59.04} & \textcolor[RGB]{64,130,109}{51.00} & \textcolor[RGB]{64,130,109}{52.38}
\\

\midrule

\multirow{2}*{\bfseries Ours (NN)} & {93.69} & 89.23 & 88.47 & \textbf{87.55} & \textbf{86.45} & 83.15 \\
& \textcolor[RGB]{64,130,109}{83.39} & \textcolor[RGB]{64,130,109}{78.95} & \textcolor[RGB]{64,130,109}{73.49} & \textcolor[RGB]{64,130,109}{71.49} & \textbf{\textcolor[RGB]{64,130,109}{71.28}} & \textcolor[RGB]{64,130,109}{70.32}
\\
\midrule
\multirow{2}*{\bfseries Ours (Proto)} & {93.69} & \textbf{89.53} & \textbf{88.50} & 87.50 & 86.20 & \textbf{84.02} \\
& \textcolor[RGB]{64,130,109}{83.39} & \textbf{\textcolor[RGB]{64,130,109}{79.84}} & \textbf{\textcolor[RGB]{64,130,109}{74.33}} & \textbf{\textcolor[RGB]{64,130,109}{72.92}} & \textcolor[RGB]{64,130,109}{70.78} & \textbf{\textcolor[RGB]{64,130,109}{72.19}}\\

\bottomrule

\end{tabular}
\caption{Detailed results of OntoNotes task order 1. The numbers in black are the micro-f1 scores and the numbers in green are the macro-f1 scores.}
\label{onto_result1}
\end{table*}

\begin{table*}[ht] 
\centering
\small
\begin{tabular}{l|c|c|c|c|c|c}
\toprule

\multirow{2}*{\bfseries Methods} & \multirow{2}*{\textbf{Step 0}}& \multirow{2}*{\textbf{Step 1}}& \multirow{2}*{\textbf{Step 2}}& \multirow{2}*{\textbf{Step 3}}& \multirow{2}*{\textbf{Step 4}}& \multirow{2}*{\textbf{Step 5}} \\ 
&&&&&\\
\midrule
\multirow{2}*{\bfseries Full Data}
 &
94.92 & 92.07 & 90.24 & 89.94 & 88.92 & 87.33\\
& \textcolor[RGB]{64,130,109}{92.85} & \textcolor[RGB]{64,130,109}{78.90} & \textcolor[RGB]{64,130,109}{78.54} & \textcolor[RGB]{64,130,109}{77.84} & \textcolor[RGB]{64,130,109}{77.97} & \textcolor[RGB]{64,130,109}{77.27}

 \\

\midrule

\multirow{2}*{\bfseries LwF} 
& 95.29 & 80.10 & 77.75 & 78.88 & 58.99 & 56.81 \\
& \textbf{\textcolor[RGB]{64,130,109}{93.19}} & \textcolor[RGB]{64,130,109}{59.83} & \textcolor[RGB]{64,130,109}{60.18} & \textcolor[RGB]{64,130,109}{61.11} & \textcolor[RGB]{64,130,109}{47.86} & \textcolor[RGB]{64,130,109}{46.95}
\\
\midrule
\multirow{2}*{\bfseries SCR} & 94.63 & 81.37 & 83.97 & 84.39 & 83.40 & 79.76 \\
& \textcolor[RGB]{64,130,109}{91.33} & \textcolor[RGB]{64,130,109}{61.60} & \textcolor[RGB]{64,130,109}{63.79} & \textcolor[RGB]{64,130,109}{63.47} & \textcolor[RGB]{64,130,109}{62.55} & \textcolor[RGB]{64,130,109}{61.78}

\\
\midrule

\multirow{2}*{\bfseries iCaRL} & \textbf{95.34} & 84.37 & 81.09 & 81.86 & 82.60 & 78.91 \\
& \textcolor[RGB]{64,130,109}{92.80} & \textcolor[RGB]{64,130,109}{69.30} & \textcolor[RGB]{64,130,109}{65.60} & \textcolor[RGB]{64,130,109}{65.79} & \textcolor[RGB]{64,130,109}{67.94} & \textcolor[RGB]{64,130,109}{66.23}
\\
\midrule


\multirow{2}*{\bfseries Con.NER}  & 94.81 & 74.22 & 72.15 & 72.68 & 73.37 & 66.37 \\
& \textcolor[RGB]{64,130,109}{92.13} & \textcolor[RGB]{64,130,109}{55.44} & \textcolor[RGB]{64,130,109}{52.84} & \textcolor[RGB]{64,130,109}{55.07} & \textcolor[RGB]{64,130,109}{53.51} & \textcolor[RGB]{64,130,109}{51.20}
\\
\midrule
\multirow{2}*{\bfseries Con.NER*} & 94.99 & 74.66 & 72.80 & 74.11 & 74.28 & 66.09 \\
& \textcolor[RGB]{64,130,109}{92.45} & \textcolor[RGB]{64,130,109}{55.63} & \textcolor[RGB]{64,130,109}{53.23} & \textcolor[RGB]{64,130,109}{56.07} & \textcolor[RGB]{64,130,109}{55.70} & \textcolor[RGB]{64,130,109}{52.45}
\\

\midrule

\multirow{2}*{\bfseries Ours (NN)} & 94.65 & \textbf{85.87} & 86.17 & \textbf{87.80} & \textbf{86.97} & \textbf{83.40} \\
& \textcolor[RGB]{64,130,109}{92.72} & \textbf{\textcolor[RGB]{64,130,109}{69.93}} & \textcolor[RGB]{64,130,109}{69.86} & \textbf{\textcolor[RGB]{64,130,109}{72.34}} & \textbf{\textcolor[RGB]{64,130,109}{72.68}} & \textbf{\textcolor[RGB]{64,130,109}{70.21}}
\\
\midrule
\multirow{2}*{\bfseries Ours (Proto)} & 94.65 & 85.33 & \textbf{86.79} & 87.13 & 86.71 & 82.75 \\
& \textcolor[RGB]{64,130,109}{92.72} & \textcolor[RGB]{64,130,109}{69.65} & \textbf{\textcolor[RGB]{64,130,109}{71.23}} & \textcolor[RGB]{64,130,109}{71.89} & \textcolor[RGB]{64,130,109}{71.72} & \textcolor[RGB]{64,130,109}{68.89}\\

\bottomrule

\end{tabular}
\caption{Detailed results of OntoNotes task order 2. The numbers in black are the micro-f1 scores and the numbers in green are the macro-f1 scores.}
\label{onto_result2}
\end{table*}

\begin{table*}[ht] 
\centering
\small
\begin{tabular}{l|c|c|c|c|c|c}
\toprule

\multirow{2}*{\bfseries Methods} & \multirow{2}*{\textbf{Step 0}}& \multirow{2}*{\textbf{Step 1}}& \multirow{2}*{\textbf{Step 2}}& \multirow{2}*{\textbf{Step 3}}& \multirow{2}*{\textbf{Step 4}}& \multirow{2}*{\textbf{Step 5}} \\ 
&&&&&\\
\midrule
\multirow{2}*{\bfseries Full Data}
 &
88.64 & 89.64 & 87.91 & 87.93 & 87.05 & 87.39\\
& \textcolor[RGB]{64,130,109}{86.15} & \textcolor[RGB]{64,130,109}{79.89} & \textcolor[RGB]{64,130,109}{76.82} & \textcolor[RGB]{64,130,109}{78.15} & \textcolor[RGB]{64,130,109}{76.80} & \textcolor[RGB]{64,130,109}{76.24}

 \\

\midrule

\multirow{2}*{\bfseries LwF} 
& 87.88 & 81.65 & 72.46 & 45.39 & 43.01 & 48.05 \\
& \textcolor[RGB]{64,130,109}{85.36} & \textcolor[RGB]{64,130,109}{65.78} & \textcolor[RGB]{64,130,109}{55.52} & \textcolor[RGB]{64,130,109}{44.41} & \textcolor[RGB]{64,130,109}{47.94} & \textcolor[RGB]{64,130,109}{46.83}
\\
\midrule
\multirow{2}*{\bfseries SCR} & 88.13 & 83.76 & 71.53 & 68.83 & 64.05 & 69.93 \\
& \textcolor[RGB]{64,130,109}{85.00} & \textcolor[RGB]{64,130,109}{68.72} & \textcolor[RGB]{64,130,109}{59.03} & \textcolor[RGB]{64,130,109}{61.20} & \textcolor[RGB]{64,130,109}{60.30} & \textcolor[RGB]{64,130,109}{58.20}

\\
\midrule

\multirow{2}*{\bfseries iCaRL} & \textbf{92.49} & 84.89 & 79.65 & 80.03 & 78.76 & 77.14 \\
& \textbf{\textcolor[RGB]{64,130,109}{87.56}} & \textcolor[RGB]{64,130,109}{72.97} & \textcolor[RGB]{64,130,109}{65.95} & \textcolor[RGB]{64,130,109}{67.17} & \textcolor[RGB]{64,130,109}{67.09} & \textcolor[RGB]{64,130,109}{65.86}
\\
\midrule


\multirow{2}*{\bfseries Con.NER}  & 88.08 & 82.24 & 66.63 & 64.16 & 59.78 & 51.94 \\
& \textcolor[RGB]{64,130,109}{85.22} & \textcolor[RGB]{64,130,109}{66.33} & \textcolor[RGB]{64,130,109}{57.53} & \textcolor[RGB]{64,130,109}{51.72} & \textcolor[RGB]{64,130,109}{48.32} & \textcolor[RGB]{64,130,109}{42.58}
\\
\midrule
\multirow{2}*{\bfseries Con.NER*} & 88.03 & 82.48 & 70.09 & 66.28 & 61.64 & 55.93 \\
& \textcolor[RGB]{64,130,109}{84.96} & \textcolor[RGB]{64,130,109}{66.32} & \textcolor[RGB]{64,130,109}{57.84} & \textcolor[RGB]{64,130,109}{55.93} & \textcolor[RGB]{64,130,109}{50.94} & \textcolor[RGB]{64,130,109}{46.77}
\\

\midrule

\multirow{2}*{\bfseries Ours (NN)} & 88.82 & \textbf{88.05} & 85.15 & \textbf{85.15} & 83.50 & 82.38 \\
& \textcolor[RGB]{64,130,109}{86.33} & \textbf{\textcolor[RGB]{64,130,109}{75.86}} & \textbf{\textcolor[RGB]{64,130,109}{71.62}} & \textbf{\textcolor[RGB]{64,130,109}{73.08}}& \textcolor[RGB]{64,130,109}{73.04} & \textcolor[RGB]{64,130,109}{69.71}
\\
\midrule
\multirow{2}*{\bfseries Ours (Proto)} & 88.82 & 87.35 & \textbf{85.31} & 84.65 & \textbf{83.99} & \textbf{82.78} \\
& \textcolor[RGB]{64,130,109}{86.33} & \textcolor[RGB]{64,130,109}{73.51} & \textcolor[RGB]{64,130,109}{70.08} & \textcolor[RGB]{64,130,109}{71.21} & \textbf{\textcolor[RGB]{64,130,109}{73.25}} & \textbf{\textcolor[RGB]{64,130,109}{69.89}}\\

\bottomrule

\end{tabular}
\caption{Detailed results of OntoNotes task order 3. The numbers in black are the micro-f1 scores and the numbers in green are the macro-f1 scores.}
\label{onto_result3}
\end{table*}

\begin{table*}[ht] 
\centering
\small
\begin{tabular}{c|c|c|c|c}
\toprule
\multirow{2}*{\bfseries Task} & \multirow{2}*{\textbf{Entity Class}}& \multirow{2}*{\textbf{\# Train}}& 
\multirow{2}*{\textbf{\# Dev}}& \multirow{2}*{\textbf{\# Test}}  \\ 
&&&&\\
\midrule

\multirow{2}*{\bfseries 1}
& 
['building-library', 'organization-showorganization', 'other-award', 
& \multirow{2}*{18435} & \multirow{2}*{2656} & \multirow{2}*{5296} \\
&'building-other', 'organization-religion', 'organization-sportsteam']&&& \\

\midrule
\multirow{2}*{\bfseries 2}
& 
['person-politician', 'art-painting', 'event-disaster', 'organization-other', 
& \multirow{2}*{18966} & \multirow{2}*{2788} & \multirow{2}*{10267} \\
&'product-weapon', 'building-hotel']&&& \\
\midrule
\multirow{2}*{\bfseries 3}
& 
['event-sportsevent', 'other-chemicalthing', 'art-writtenart',
& \multirow{2}*{11973} & \multirow{2}*{1652} & \multirow{2}*{13055} \\
&'product-game', 'location-mountain', 'other-livingthing']&&& \\
\midrule
\multirow{2}*{\bfseries 4}
&  
['location-island', 'person-scholar', 'building-restaurant',
& \multirow{2}*{9448} & \multirow{2}*{1326} & \multirow{2}*{15178} \\
&'other-astronomything', 'building-airport', 'product-other']&&& \\
\midrule
\multirow{2}*{\bfseries 5}
&  
['location-road/railway/highway/transit', 'other-educationaldegree',  
& \multirow{2}*{10295} & \multirow{2}*{1477} & \multirow{2}*{17254} \\
&'building-sportsfacility', 'event-election', 'person-actor', 'art-film']&&& \\
\midrule
\multirow{2}*{\bfseries 6}
&  
['location-other', 'product-ship', 'organization-politicalparty',  
& \multirow{2}*{47648} & \multirow{2}*{6941} & \multirow{2}*{24429} \\
&'person-soldier', 'location-GPE', 'other-god']&&& \\
\midrule
\multirow{2}*{\bfseries 7}
&  
['event-attack/battle/war/militaryconflict', 'organization-sportsleague',  
& \multirow{2}*{13237} & \multirow{2}*{1852} & \multirow{2}*{25631} \\
&'building-theater', 'organization-education', 'product-train', 'other-medical']&&& \\
\midrule
\multirow{2}*{\bfseries 8}
&  
['event-protest', 'person-other', 'product-car', 'art-other',  
& \multirow{2}*{30899} & \multirow{2}*{4416} & \multirow{2}*{29014} \\
&'organization-company', 'other-disease']&&& \\
\midrule
\multirow{2}*{\bfseries 9}
&  
['other-biologything', 'person-artist/author', 'location-bodiesofwater', 
& \multirow{2}*{21794} & \multirow{2}*{3114} & \multirow{2}*{31036} \\
&'art-broadcastprogram', 'other-language', 'person-athlete']&&& \\
\midrule
\multirow{2}*{\bfseries 10}
&  
['product-airplane', 'art-music', 'product-software', 
& \multirow{2}*{11963} & \multirow{2}*{1706} & \multirow{2}*{31874} \\
&'event-other', 'location-park', 'organization-media/newspaper']&&& \\
\midrule
\multirow{2}*{\bfseries 11}
&  
['other-currency', 'person-director', 'building-hospital', 'other-law', 
& \multirow{2}*{9787} & \multirow{2}*{1443} & \multirow{2}*{32565} \\
&'organization-government/governmentagency', 'product-food']&&& \\
\bottomrule

\end{tabular}
\caption{Details of Few-NERD task order 1.}
\label{nerd_detail1}
\end{table*}

\begin{table*}[ht] 
\centering
\small
\begin{tabular}{c|c|c|c|c}
\toprule
\multirow{2}*{\bfseries Task} & \multirow{2}*{\textbf{Entity Class}}& \multirow{2}*{\textbf{\# Train}}& 
\multirow{2}*{\textbf{\# Dev}}& \multirow{2}*{\textbf{\# Test}}  \\ 
&&&&\\
\midrule

\multirow{2}*{\bfseries 1}
& 
['location-GPE', 'event-sportsevent', 'organization-showorganization', 
& \multirow{2}*{48730} & \multirow{2}*{7060} & \multirow{2}*{13963} \\
&'event-attack/battle/war/militaryconflict', 'art-other', 'product-car']&&& \\

\midrule
\multirow{2}*{\bfseries 2}
& 
['location-bodiesofwater', 'person-scholar', 'person-artist/author',  
& \multirow{2}*{21523} & \multirow{2}*{3183} & \multirow{2}*{17477} \\
&'person-politician', 'other-livingthing', 'product-airplane']&&& \\
\midrule
\multirow{2}*{\bfseries 3}
& 
['product-other', 'art-music', 'location-island', 
& \multirow{2}*{15257} & \multirow{2}*{2169} & \multirow{2}*{19805} \\
&'person-athlete', 'building-airport', 'building-hotel']&&& \\
\midrule
\multirow{2}*{\bfseries 4}
&  
['person-soldier', 'event-other', 'product-software', 'event-election', 
& \multirow{2}*{17967} & \multirow{2}*{2531} & \multirow{2}*{22192} \\
&'organization-other', 'organization-politicalparty']&&& \\
\midrule
\multirow{2}*{\bfseries 5}
&  
['other-award', 'art-film', 'organization-government/governmentagency', 
& \multirow{2}*{12258} & \multirow{2}*{1792} & \multirow{2}*{23841} \\
&'other-astronomything', 'person-actor', 'person-director']&&& \\
\midrule
\multirow{2}*{\bfseries 6}
&  
['event-protest', 'building-library', 'art-broadcastprogram', 
& \multirow{2}*{11191} & \multirow{2}*{1620} & \multirow{2}*{25125} \\
&'other-educationaldegree', 'organization-sportsleague', 'location-other']&&& \\
\midrule
\multirow{2}*{\bfseries 7}
&  
['product-game', 'event-disaster', 'product-train', 
& \multirow{2}*{11243} & \multirow{2}*{1578} & \multirow{2}*{26473} \\
&'building-other', 'other-disease', 'building-hospital']&&& \\
\midrule
\multirow{2}*{\bfseries 8}
&  
['product-ship', 'other-currency', 'art-painting', 
& \multirow{2}*{28035} & \multirow{2}*{4055} & \multirow{2}*{29113} \\
&'product-weapon', 'organization-sportsteam', 'person-other']&&& \\
\midrule
\multirow{2}*{\bfseries 9}
&  
['other-god', 'art-writtenart', 'other-chemicalthing', 
& \multirow{2}*{11390} & \multirow{2}*{1631} & \multirow{2}*{30302} \\
&'organization-education', 'other-medical', 'building-restaurant']&&& \\
\midrule
\multirow{2}*{\bfseries 10}
&  
['building-sportsfacility', 'building-theater', 'organization-company', 
& \multirow{2}*{16806} & \multirow{2}*{2333} & \multirow{2}*{31810} \\
&'other-biologything', 'organization-religion', 'other-law']&&& \\
\midrule
\multirow{2}*{\bfseries 11}
&  
['location-mountain', 'location-road/railway/highway/transit', 
& \multirow{2}*{11256} & \multirow{2}*{1594} & \multirow{2}*{32565} \\
&'organization-media/newspaper', 'location-park', 'product-food', 'other-language']&&& \\
\bottomrule

\end{tabular}
\caption{Details of Few-NERD task order 2.}
\label{nerd_detail2}
\end{table*}

\begin{table*}[ht] 
\centering
\small
\begin{tabular}{c|c|c|c|c}
\toprule
\multirow{2}*{\bfseries Task} & \multirow{2}*{\textbf{Entity Class}}& \multirow{2}*{\textbf{\# Train}}& 
\multirow{2}*{\textbf{\# Dev}}& \multirow{2}*{\textbf{\# Test}}  \\ 
&&&&\\
\midrule

\multirow{2}*{\bfseries 1}
& 
['organization-other', 'art-film', 'product-weapon',  
& \multirow{2}*{24344} & \multirow{2}*{3377} & \multirow{2}*{6860} \\
&'building-sportsfacility', 'person-soldier', 'organization-company']]&&& \\

\midrule
\multirow{2}*{\bfseries 2}
& 
['person-actor', 'product-other', 'person-athlete',  
& \multirow{2}*{18381} & \multirow{2}*{2603} & \multirow{2}*{11276} \\
&'building-theater', 'organization-media/newspaper', 'event-other']&&& \\
\midrule
\multirow{2}*{\bfseries 3}
& 
['event-attack/battle/war/militaryconflict',  'organization-showorganization'
& \multirow{2}*{102711} & \multirow{2}*{1504} & \multirow{2}*{13318} \\
& 'other-livingthing', 'other-language', 'art-broadcastprogram', 'product-ship']&&& \\
\midrule
\multirow{2}*{\bfseries 4}
&  
['other-award', 'location-road/railway/highway/transit','event-election', 
& \multirow{2}*{25684} & \multirow{2}*{3748} & \multirow{2}*{18354} \\
& 'event-protest', 'person-other', 'art-painting']&&& \\
\midrule
\multirow{2}*{\bfseries 5}
&  
['other-medical', 'other-chemicalthing', 'product-airplane', 
& \multirow{2}*{14767} & \multirow{2}*{2148} & \multirow{2}*{20948} \\
&'art-music', 'organization-education', 'location-bodiesofwater'] &&& \\
\midrule
\multirow{2}*{\bfseries 6}
&  
['other-astronomything', 'building-library', 'organization-sportsteam', 
& \multirow{2}*{15831} & \multirow{2}*{2302} & \multirow{2}*{23476} \\
&'product-food', 'building-restaurant', 'person-politician']&&& \\
\midrule
\multirow{2}*{\bfseries 7}
&  
['other-biologything', 'location-mountain', 'location-other', 
& \multirow{2}*{12075} & \multirow{2}*{1723} & \multirow{2}*{25509} \\
&'building-airport', 'other-currency', 'other-educationaldegree']&&& \\
\midrule
\multirow{2}*{\bfseries 8}
&  
['organization-politicalparty', 'product-car', 'building-hotel', 
& \multirow{2}*{14431} & \multirow{2}*{2089} & \multirow{2}*{27139} \\
&'location-island', 'person-artist/author', 'other-law']&&& \\
\midrule
\multirow{2}*{\bfseries 9}
&  
['product-train', 'organization-government/governmentagency', 'other-disease', 
& \multirow{2}*{9792} & \multirow{2}*{1388} & \multirow{2}*{28142} \\
&'person-director', 'location-park', 'event-disaster']&&& \\
\midrule
\multirow{2}*{\bfseries 10}
&  
['art-writtenart', 'other-god', 'art-other', 'organization-sportsleague', 
& \multirow{2}*{47410} & \multirow{2}*{6902} & \multirow{2}*{31564} \\
&'organization-religion', 'location-GPE']&&& \\
\midrule
\multirow{2}*{\bfseries 11}
&  
['product-game', 'product-software', 'person-scholar', 
& \multirow{2}*{15549} & \multirow{2}*{2157} & \multirow{2}*{32565} \\
&'event-sportsevent', 'building-hospital', 'building-other']&&& \\
\bottomrule

\end{tabular}
\caption{Details of Few-NERD task order 3.}
\label{nerd_detail3}
\end{table*}

\begin{table*}[ht] 
\centering
\small
\begin{tabular}{c|c|c|c|c}
\toprule
\multirow{2}*{\bfseries Task} & \multirow{2}*{\textbf{Entity Class}}& \multirow{2}*{\textbf{\# Train}}& 
\multirow{2}*{\textbf{\# Dev}}& \multirow{2}*{\textbf{\# Test}}  \\ 
&&&&\\
\midrule

\bfseries 1
& 
['PRODUCT', 'GPE', 'CARDINAL']
& 15119 & 2149 & 2124 \\

\midrule
\bfseries 2
& 
['QUANTITY', 'DATE', 'LANGUAGE']
& 9561 & 1335 & 2883 \\
\midrule
\bfseries 3
& 
['PERSON', 'LAW', 'LOC']
& 13424 & 1725 & 3852 \\

\midrule
\bfseries 4
&  
['ORDINAL', 'PERCENT', 'EVENT']
& 3259 & 460 & 4002 \\

\midrule
\bfseries 5
&  
['NORP', 'FAC', 'TIME']
& 6913 & 949 & 4291 \\

\midrule
\bfseries 6
&  
['MONEY', 'WORK\_OF\_ART','ORG'] 
& 11286 & 1480 & 4624 \\

\bottomrule

\end{tabular}
\caption{Details of OntoNotes 5.0 task order 1.}
\label{onto_detail1}
\end{table*}

\begin{table*}[ht] 
\centering
\small
\begin{tabular}{c|c|c|c|c}
\toprule
\multirow{2}*{\bfseries Task} & \multirow{2}*{\textbf{Entity Class}}& \multirow{2}*{\textbf{\# Train}}& 
\multirow{2}*{\textbf{\# Dev}}& \multirow{2}*{\textbf{\# Test}}  \\ 
&&&&\\
\midrule

\bfseries 1
& 
['ORDINAL', 'PERSON', 'PERCENT']
& 14323 & 1814 & 1919 \\

\midrule
\bfseries 2
& 
['WORK\_OF\_ART', 'PRODUCT', 'LAW']
& 1634 & 229 & 2073 \\
\midrule
\bfseries 3
& 
['CARDINAL', 'EVENT', 'QUANTITY']
& 6786 & 904 & 2711 \\

\midrule
\bfseries 4
&  
['GPE', 'MONEY', 'TIME']
& 12823 & 1887 & 3718 \\

\midrule
\bfseries 5
&  
['NORP', 'LANGUAGE', 'DATE']
& 13090 & 1820 & 4318 \\

\midrule
\bfseries 6
&  
['LOC', 'FAC', 'ORG'] 
& 11127 & 1500 & 4624 \\

\bottomrule

\end{tabular}
\caption{Details of OntoNotes 5.0 task order 2.}
\label{onto_detail2}
\end{table*}

\begin{table*}[ht] 
\centering
\small
\begin{tabular}{c|c|c|c|c}
\toprule
\multirow{2}*{\bfseries Task} & \multirow{2}*{\textbf{Entity Class}}& \multirow{2}*{\textbf{\# Train}}& 
\multirow{2}*{\textbf{\# Dev}}& \multirow{2}*{\textbf{\# Test}}  \\ 
&&&&\\
\midrule

\bfseries 1
& 
['ORG', 'CARDINAL', 'QUANTITY']
& 14284 & 1898 & 1867 \\

\midrule
\bfseries 2
& 
['LAW', 'FAC', 'GPE']
& 11335 & 1692 & 2877 \\
\midrule
\bfseries 3
& 
['DATE', 'LANGUAGE', 'WORK\_OF\_ART']
& 9791 & 1386 & 3522 \\

\midrule
\bfseries 4
&  
['PERCENT', 'NORP', 'EVENT']
& 6927 & 927 & 3815 \\

\midrule
\bfseries 5
&  
['ORDINAL', 'TIME', 'MONEY']
& 4327 & 596 & 3979 \\

\midrule
\bfseries 6
&  
[ 'PERSON', 'LOC', 'PRODUCT'] 
& 13663 & 1747 & 4624 \\

\bottomrule

\end{tabular}
\caption{Details of OntoNotes 5.0 task order 3.}
\label{onto_detail3}
\end{table*}

\end{document}